\AtBeginDocument{%
  \paperwidth=\dimexpr
    1in + \oddsidemargin
    + \textwidth
    + 1in + \oddsidemargin
  \relax
  \paperheight=\dimexpr
    1in + \topmargin
    + \headheight + \headsep
    + \textheight
    + 1in + \topmargin
  \relax
  \usepackage[pass]{geometry}\relax
}

\RequirePackage{fix-cm}
\documentclass[smallextended]{svjour3}       
\smartqed  
\usepackage{siunitx}

\usepackage{afterpage}

\usepackage{multirow}
\usepackage{graphicx}
\usepackage{hhline}
\usepackage{longtable}
\usepackage{rotating}

\usepackage[T1]{fontenc}

\usepackage{marvosym}
\usepackage{xcolor}
\usepackage{float}
\definecolor{darkgreen}{RGB}{15, 127, 4}

\usepackage{enumitem}   

\begin{document}

\newpage
\setcounter{page}{1}

\title{
Spatiotemporal Data Mining: A Survey on Challenges and Open Problems
%
}


\author{Ali Hamdi \textsuperscript{1} . \thanks{\textsuperscript{1} School of Computing Technologies, RMIT University, Australia}
Khaled Shaban \textsuperscript{2} \thanks{\textsuperscript{2} Department of Computer Science and Engineering, Qatar University, Qatar} . \\
Abdelkarim Erradi \textsuperscript{2} . Amr Mohamed \textsuperscript{2}\\
Shakila Khan Rumi \textsuperscript{1}.
Flora Salim \textsuperscript{1}
}

\authorrunning{Spatiotemporal Data Mining: A Survey on Challenges and Open Problems} 

\institute{ \Letter Ali Hamdi \at \email{ali.ali@rmit.edu.au}}

\date{Received: 24 Jun 2019 / Accepted: 29 Mar 2021}
\maketitle

\begin{abstract}\label{abstract}
Spatiotemporal data mining (STDM) discovers useful patterns from the dynamic interplay between space and time. 
Several available surveys capture STDM advances and report a wealth of important progress in this field. However, STDM challenges and problems are not thoroughly discussed and presented in articles of their own. {We attempt to fill this gap by providing a comprehensive literature survey on state-of-the-art advances in STDM.}
We describe the challenging issues and their causes and open gaps of multiple STDM directions and aspects. Specifically, we investigate the challenging issues in regards to spatiotemporal relationships, interdisciplinarity, discretisation, and data characteristics. Moreover, we discuss the limitations in the literature and open research problems related to spatiotemporal data representations, modelling and visualisation, and comprehensiveness of approaches. We explain issues related to STDM tasks of classification, clustering, hotspot detection, association and pattern mining, outlier detection, visualisation, visual analytics, and computer vision tasks. We also highlight STDM issues related to multiple applications including crime and public safety, traffic and transportation, earth and environment monitoring, epidemiology, social media, and Internet of Things. 

\keywords{Spatial, Spatiotemporal, Data Mining, Challenges Issues, Research Problems}
\end{abstract}

\setcounter{figure}{0}

\section{Introduction}\label{introduction}
There has been an increase in the research of Spatiotemporal Data Mining (STDM) due to growing availability of geo-referenced and temporal data and also due to the complexity and poor performance when applying classical data mining methods \cite{Shekhar2015perspective,RN266}. 
{
Large amounts of spatiotemporal data are being generated and captured through systems that record sequential observations of remote sensing, mobility, wearable devices, and social media. Spatiotemporal data represent different phenomena ranging from micro-scale of DNA and cell evolution, to global ones, e.g., climate change \cite{Chaowei2019Big}. The wide-availability of user-generated data via the social media platforms offers great opportunities to understand people needs, thoughts, and sentiments toward specific topics, products, or services \cite{hamdi2018clasenti}. Medical sensory devices observe different activities at various locations of the human body over specific time ranges.
STDM proposes new methods to handle such data through advanced predictive and descriptive tasks such as classification and clustering to work best with space and time referenced data. 
STDM methods are concerned with relationships and dependencies among different measurements. These relationships are complex, implicit, and dynamically changing.
To a large extent, classical data mining assumes that data are independent and identically distributed (i.i.d.). On the contrast, spatiotemporal data do not follow this assumption and STDM methods aim to capture the autocorrelation among different events or data points that are interdisciplinary in nature, i.e, data from multiple domains that may require the utilisation of various mining tasks.
\newline
\newline
Spatiotemporal data comprise spatial and temporal representations. They include three distinct types of attributes, namely, non-spatiotemporal, spatial and temporal attributes \cite{RN14t}. The non-spatiotemporal attributes represent non-contextual features of objects. Spatial attributes define the locations, extents, and shapes of the objects. Temporal attributes are timestamps and durations of processes denoting spatial object (vector) or field (raster layers). For example, air pollution spatiotemporal data have non-spatiotemporal attributes such as air pollution levels or station names, spatial coordinates of the location where the measurements are taken and temporal timestamps associated with the collected measurements.
Spatiotemporal data types can also be categorised based on their collection nature to discrete or continuous observations. Events and data trajectories are examples of spatiotemporal discrete data types, while continuous data types include point reference and raster data.
Spatiotemporal event data constitute discrete events that happen at geo-locations and times such as traffic accidents and crime incidents. Figure \ref{ST_types} (a) denotes spatiotemporal events of three different types. Each type is presented in different colour and shape, e.g., red circles are for events of one type and each event is annotated with the its location and time (location $l_1$ in time $t_1$).
Trajectory data contain sequences of spatiotemporal instances that trace motions of objects in geographical spaces overtimes. Trajectories are usually represented by a series of chronologically ordered points which consist of spatial coordinates and timestamps \cite{zheng2015trajectory}. For instance, a vehicle trajectory between two locations is a set of consecutive points of space and time. Figure \ref{ST_types} (b) illustrates trajectories of three coloured objects between locations ($l_1$ and $l_n$) at times ($t_1$ and $t_2$). 
Point reference data measure continuous spatiotemporal fields at moving spatiotemporal reference sites. For example, spatiotemporal point references can be utilised to measure surface temperature using moving balloons. Figure \ref{ST_types} (c and d) show spatiotemporal point reference data at different locations (black squares) at timestamps ($t_1$ and $t_2$).
Raster data represent measurements of spatiotemporal fields at fixed cells in  grids such as activities in fMRI brain scans. Figure \ref{ST_types} (e and f) show spatiotemporal raster data of regular grid at time ($t_1$ and $t_2$).
These different types of spatiotemporal data are associated with different research challenges \cite{RN357}. Classical data mining approaches are not designed to handle such data. 
\begin{figure}[]
\centering
\includegraphics[width=0.95\textwidth, angle=0]{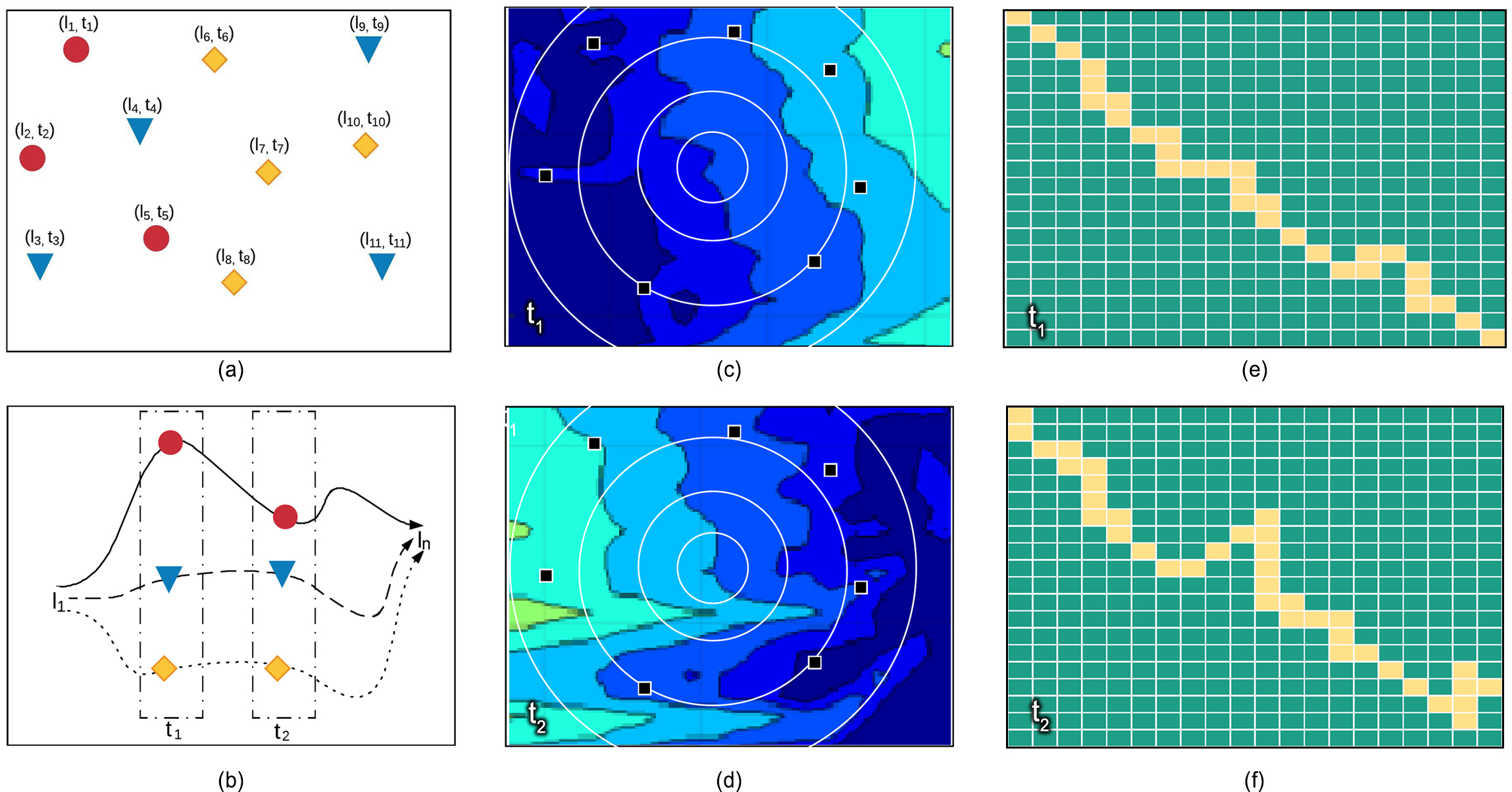}
\caption{{Spatiotemporal data types. (a) spatiotemporal events of different types at different locations and timestamps. (b) spatiotemporal trajectories between locations ($l_1$ and $l_n$) at time ($t_1$ and $t_2$). (c and d) spatiotemporal point reference data at different locations at timestamps ($t_1$ and $t_2$). (e and f) spatiotemporal raster data of regular grid at time ($t_1$ and $t_2$).}}
\label{ST_types}
\end{figure}
}

This paper consolidates the current state of the challenges associated with the STDM tasks and applications. 
{There have been several survey articles that reviewed work related to STDM, each of which discussed the literature from different perspectives such as spatial databases \cite{koperski1996spatial}, spatial patterns \cite{RN172}, spatiotemporal cluster analysis \cite{RN173}, urban concepts and applications \cite{Zheng2014Urban}, big data analytics \cite{yang2019big}, big climate data analytics \cite{hu2018climatespark}, and outliers detection \cite{RN174,meng2018overview}.} 
\cite{RN357} surveyed STDM methods and techniques according to main spatiotemporal problems of clustering, predictive learning, change detection, frequent pattern mining, anomaly detection, and relationship mining. \cite{pei2020big} reviewed the big geo-data mining objectives and issues in terms of human behaviour and distributions of geographical patterns.
The survey by \cite{Shekhar2015perspective} divided prior surveys in the literature into two types; articles with statistical foundations \cite{RN169,koperski1996spatial,RN171}, and others without that \cite{RN174,RN173,RN172}. \cite{wang2019deep} surveyed the utilised deep learning methods in STDM based on the data types, tasks and deep learning models. The authors also presented the utilisation of deep learning methods in various applications. \cite{Zheng2014Urban} surveyed the concepts and applications of urban computing and discussed their computing challenges. The work in \cite{shi2018machine} presented a review of machine learning methods for STDM sequence forecasting related problem. They focused on moving point cloud, regular grid, and irregular grids.
Due to the fast pace of advances in STDM, there is a continuous need for up-to-date surveys. 
Moreover, to the best of our knowledge, STDM  challenges and problems are not thoroughly discussed and presented in articles of their own. Specifically, none of these existing researches paid their focus on the general challenging issues in terms of relationships, data, natures and limitations of STDM research or the challenges related STDM tasks and applications. Our survey attempts to fill this gap providing a comprehensive literature survey on state-of-the-art advances in STDM. Unlike existing survey papers, we review previous works and describe STDM challenges and their causes as well as issues related to selected applications and tasks. 

\begin{figure}[]
\centering
\includegraphics[width=0.99\textwidth, angle=0]{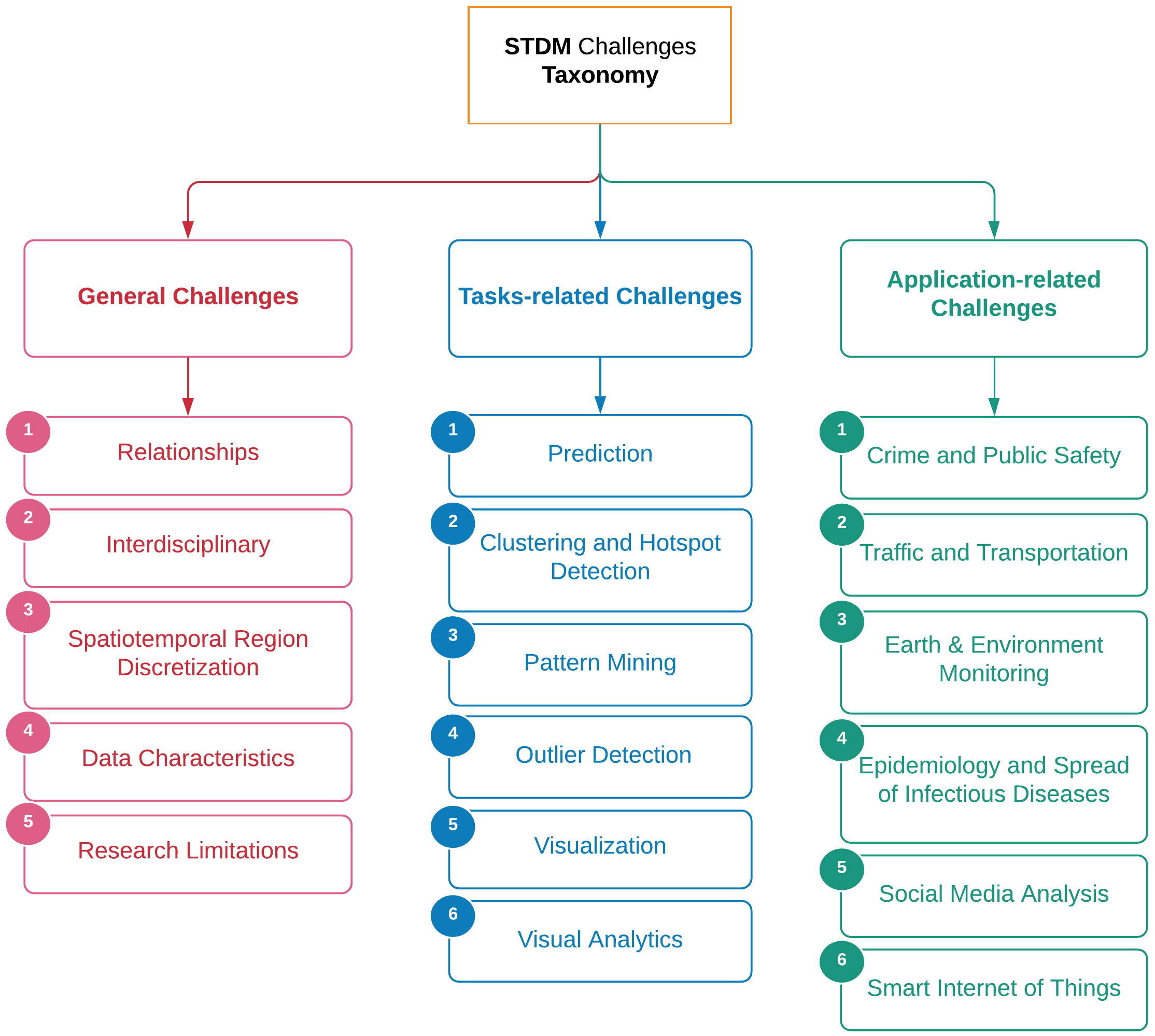}
\caption{{A taxonomy of the proposed STDM challenges structure. The survey is designed to cover the STDM related challenges from three different perspectives. We propose to investigate the general challenges that affect the STDM in terms of relationships, data, natures and limitations of research. Then, we discuss the STDM tasks and applications focusing on their related challenges.}}
\label{taxonomy}
\end{figure}

Figure \ref{taxonomy} shows a taxonomy of the proposed structure for reviewing the STDM challenges. The taxonomy highlights the survey main sections and their sub-sections. The survey is designed to cover the STDM related challenges from three different perspectives. We start the survey by defining the general challenging issues in terms of relationships, data, natures and limitations of research. Then, we discuss the STDM tasks and applications focusing on their related challenges. Finally, we conclude the survey with a mapping table and a discussion to connect the general challenges with the tasks and applications sections.

The rest of the paper is organised as follows. The research methodology on how we conducted this survey is presented in Section \ref{methodology}. Section \ref{general} discusses general STDM challenges and their causes. Section \ref{tasks} covers STDM tasks and their related challenges. Section \ref{domains} introduces STDM applications and related challenges in them. Section \ref{summary} summarises the survey and the integration between the general challenges and STDM related tasks and applications . Section \ref{conclusion} highlights key conclusions and directions for future work.

\section{Survey Methodology}\label{methodology}
\begin{figure}[]
\centering
\includegraphics[width=0.95\textwidth, angle=0]{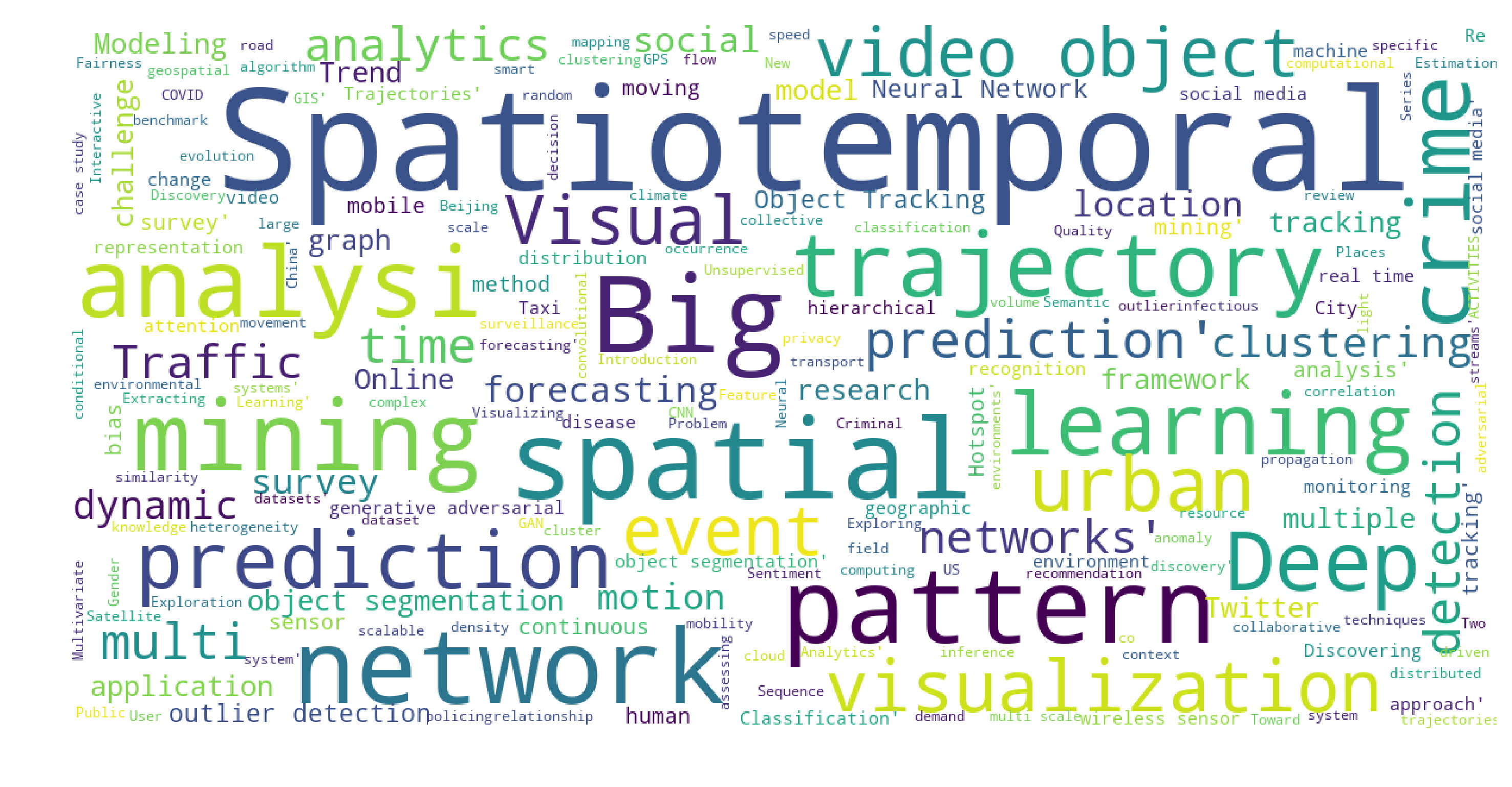}
\caption{{A word-cloud visualisation of the most frequent used search keywords.}}
\label{words}
\end{figure}

\begin{figure}[]
\centering
\includegraphics[width=0.75\textwidth, angle=0]{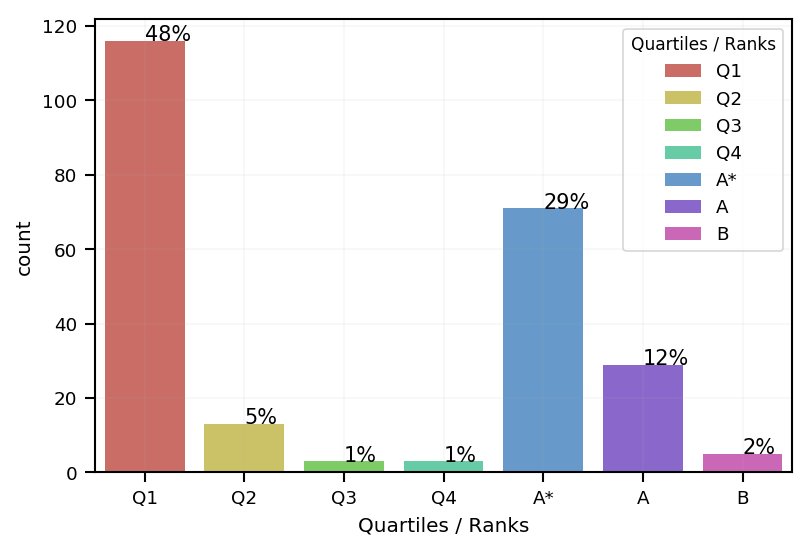}
\caption{{Related work distributions for journal articles and conference proceedings.}}
\label{ranks}
\end{figure}

\begin{figure}[]
\centering
\includegraphics[width=0.75\textwidth, angle=0]{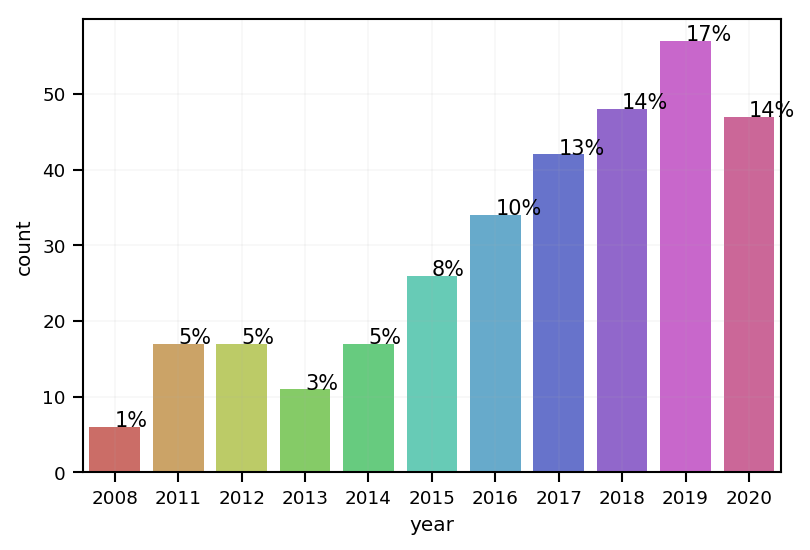}
\caption{{Related work distributions for years from 2011 to 2020.}}
\label{years}
\end{figure}

We designed our survey to focus on STDM challenges and research problems, as shown in Figure \ref{taxonomy}. We built a comprehensive set of STDM challenges, as shown in Figure \ref{challenges}. This list of challenges is accompanied with their root causes. We started extracting these challenges and causes from existing STDM surveys in addition to our knowledge in the area. We then extended the challenges set in terms of STDM tasks and applications. For example, an STDM previous survey may list several challenges that are relevant to its scope, e.g., STDM visualisation. We add these challenges to our survey by extending their definitions and searching for their related work. The most frequent search keywords are visualised in a word cloud in Figure \ref{words}.
We included $342$ STDM related work in our survey. These citations are from different publication types, including journal articles, conference proceedings, books, book chapters, and theses. Figure \ref{ranks} compares between the different ranks and quartiles of the indexed journal articles and conference proceedings. 
It shows that Q1 journals are the most cited with $48$ percent, followed by A* ranked conferences with $29$ percent. The ranks and quartiles are calculated at the Scimago Institutions Rankings (SJR) \footnote{https://www.scimagojr.com/} and Computing Research \& Education (CORE) \footnote{http://portal.core.edu.au/conf-ranks/} in December 2020. We focused the search process on high ranked journals such as: 
IEEE Transactions on Big Data, 
ACM Transactions on Intelligent Systems and Technology, Cartography and Geographic Information Science, 
IEEE Transactions on Knowledge and Data Engineering, 
Neurocomputing, and
Transactions in GIS. 
We also covered a set of highly influential conferences such as
ACM SIGKDD International Conference on Knowledge Discovery and Data Mining,
Proceedings of the IEEE Conference on Computer Vision and Pattern Recognition (CVPR),
IEEE International Conference on Data Engineering (ICDE),
IEEE Visualization Conference, 
AAAI Conference on Artificial Intelligence,
International Joint Conference on Artificial Intelligence (IJCAI), and
international conference on Ubiquitous computing.
The majority of the cited papers, 85 percent, are published in the last ten years. Figure \ref{years} shows the distributions for the citation of each year between 2011 and 2020.

\section{General STDM Challenges and Research Gaps}\label{general}
\begin{figure}[]
\centering
\includegraphics[width=.95\textwidth, angle=0]{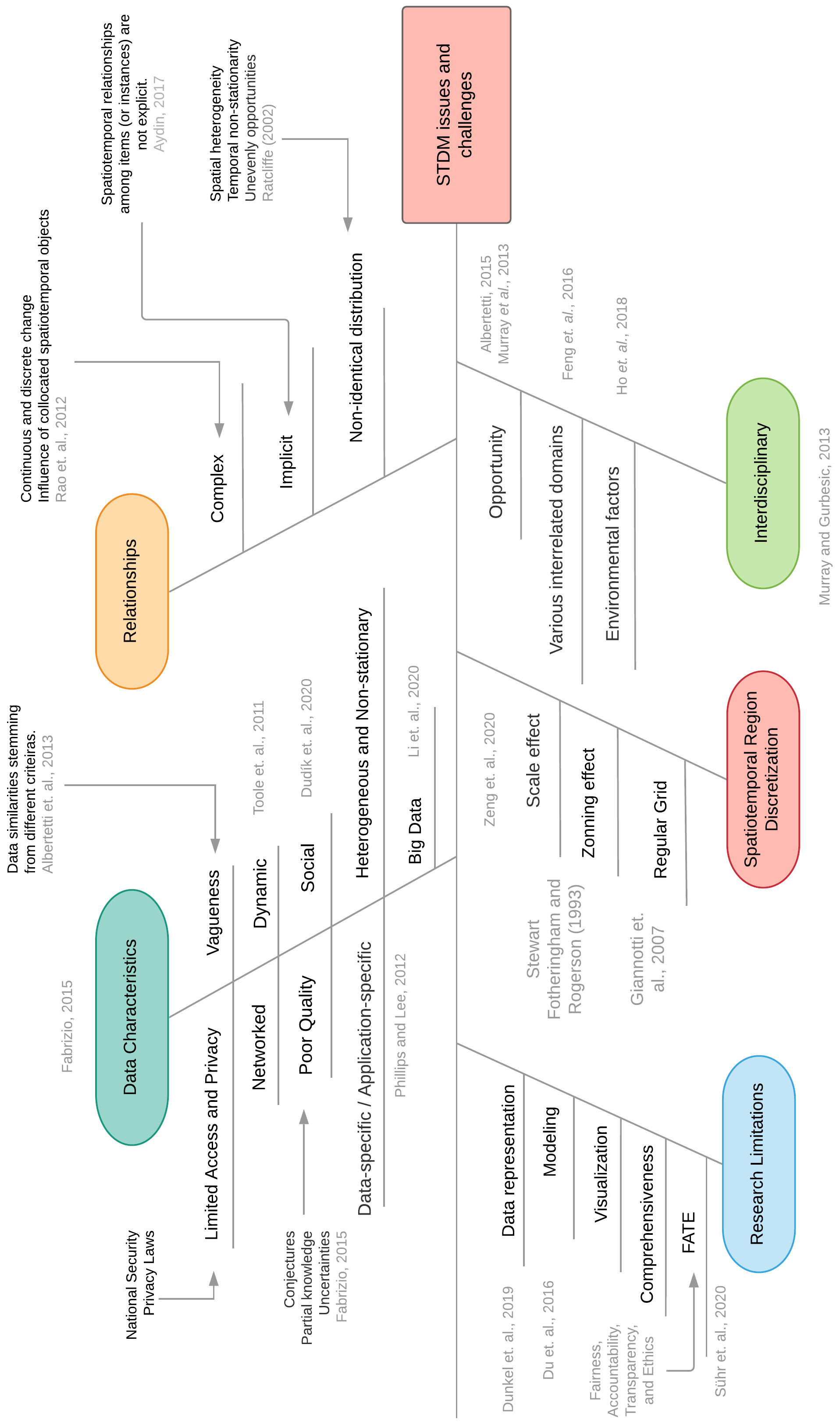}
\caption{Cause-and-effect diagram of STDM general challenges. The figure shows a taxonomy of STDM challenging issues.}
\label{challenges}
\end{figure}

There are various factors causing difficulties in STDM. We identify and list them as follows:
\begin{enumerate}
\item Spatiotemporal objects relationships that are complex and implicit. 
\item STDM requires interdisciplinary effort and integration of various heterogeneous datasets and multiple data mining algorithms.
\item Spatiotemporal region discretisation problem caused by the scale and the zoning effects on the data mining results.
\item Data characteristics such as heterogeneity and dynamicity.
\item Further Efforts Needed in STDM for data representations, advanced modelling, visualisation, and comprehensiveness.
\end{enumerate}
Figure \ref{challenges} presents these general challenges using a cause-and-effect diagram while citing related literature that explains further the used terminologies. In the next subsections, each of these challenges is discussed.

\subsection{Spatiotemporal Relationships}
Spatiotemporal objects that exist in one area or during the same time and share similar characteristics are often related. Finding relationships between objects is helpful in different tasks such as spatiotemporal hotspot prediction \cite{RN138}. However, discovering valuable relationships from spatiotemporal data is more challenging compared to traditional numerical and categorical data because of the complex data characteristics. The next sub-sections describe three of these characteristics of spatiotemporal relationships, namely; complexity, implicitness, and non-identical distributions.

\subsubsection{Complexity}
The complexity of spatiotemporal relationships poses difficulty to extracting spatiotemporal patterns \cite{Shekhar2015perspective,Shen_2019_CVPR}. \cite{RN52} stated that this complexity stems from the fact that spatiotemporal data are discrete representations of what are, in reality, continuous in space and time. For example, traffic sensing devices that are fixed in roads capture data of moving vehicles in certain locations while these vehicles are continuously moving. Moreover, co-located spatiotemporal objects influence each other and hinder the detection of relationships. {In other words, the pattern of a moving object might be affected by nearby objects such as a car's direction, speed and acceleration are influenced by other cars around it.} 

\subsubsection{Implicitness}
\begin{figure}[]
\centering
\includegraphics[width=12cm]{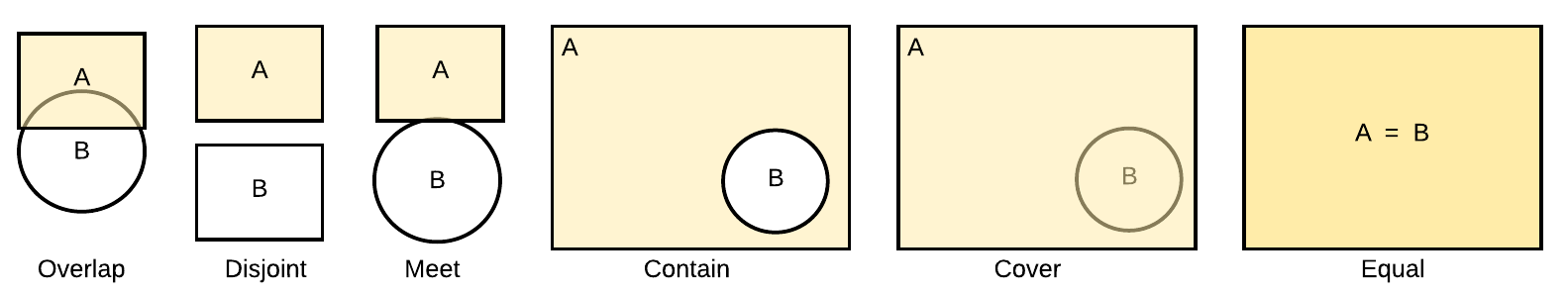}
\caption{Examples of topological relationships between two areas.}
\label{relationships}
\end{figure}

Non-spatiotemporal data have explicit relationships represented through arithmetic relations, such as ordering, instance-of, subclass-of, and member-of. On the contrary, relationships between spatiotemporal objects are implicit \cite{RN172}. Spatial relationships are built based on qualities or feature such as distance, volume, size and time. These relationships can occur among points, lines, regions or a mixture of them. For instance, Figure \ref{relationships} shows that topological relationships between two regions, include, disjoint, overlap, contains, covers, meet, equal, inside and covered-by \cite{RN355}. 
%
A spatiotemporal point can co-locate with another point. A line in a spatiotemporal environment can intersect, overlap, touch or be within another line or spatiotemporal area.
%
For example, in the case of migrant birds, flying birds can be described as a complex network of multiple spatiotemporal lines. The work in \cite{RN356} analysed a daily temporal resolution for migration trajectories of 118 migratory bird species from 2002 to 2014. In order to address this issue, the spatiotemporal relationships can be transformed into traditional relationships mined using classical data mining methods. However, this process causes information losses which inevitably preclude detecting subtle relationships. For example, in the case of traffic monitoring, using distributed sensing systems capture micro-scale sensing data for the whole sensed area; while using fixed traffic sensors can only collect transaction data for anonymous moving objects' speed, direction and acceleration. The former represents the actual, but implicit, movement patterns. {The latter aggregates the sensing data. This data aggregations leads to loss of data about tracking the relationships between them.}

\subsubsection{Non-independent and Non-identical Distribution}
Spatiotemporal data objects have positive autocorrelation or dependency. Nearby things in space and time tend to be related and more similar than distant things. In moving cars, for example, there are many dependent variables such as location, direction, connectivity and temporal attributes \cite{RN171}. For instance, if there is heavy traffic in an intersection at 4 pm, the chances are that there will be some traffic at 4.01 pm as well. 
Moreover, opportunity contributes more to the probability that a spatiotemporal pattern to occur \cite{RN139}. For example, crime often occurs when a criminal and a victim are found in the same location and time.
This is an autocorrelation as opposed to classical data that are independent and identically distributed. The autocorrelation between spatiotemporal objects degrades the performance of data mining algorithms \cite{RN175}. Additionally, measuring the spatiotemporal autocorrelation in large datasets is computationally expensive.
\\
Identifying spatiotemporal distribution characteristics is useful for patterns discovery. It is also important for detecting regularly repeating relationships between spatiotemporal objects \cite{RN184}. Spatiotemporal distribution was discussed in various studies in different domains such as phenology, geology, ecotoxicology, and criminology. Such studies reported that spatiotemporal data tend to have a non-identical distribution across space (spatial heterogeneity) and over time (temporal non-stationarity). A spatiotemporal dataset may have geographical regions and temporal periods with distinguishable distributions \cite{RN150}. For example, drivers' behaviours are varying concerning location and time. Driving seems to be safer in quiet areas and night hours. While it is expected to be dangerous in crowded areas, especially during rush hours.
Moreover, spatiotemporal data has skewed distribution in different locations of a city, e.g., a city downtown may have high-volume data than the other suburbs. This issue can affect the performance of STDM tasks.

\subsection{Interdisciplinary and Combined Data Mining}
STDM requires interdisciplinary efforts and ever-expanding knowledge from different domains. 
%
For instance, spatiotemporal crime data analysis requires large-scale macro datasets analysis for socio-economy, socio-psychology, culture and demography \cite{RN147}, and micro-environmental datasets such as interurban structure, distance, density clusters and tactics to crimes \cite{RN283}.
%
Other factors also have been investigated, such as globalisation and social and demographic change.  
\begin{figure}[]
\centering
\includegraphics[width=7cm]{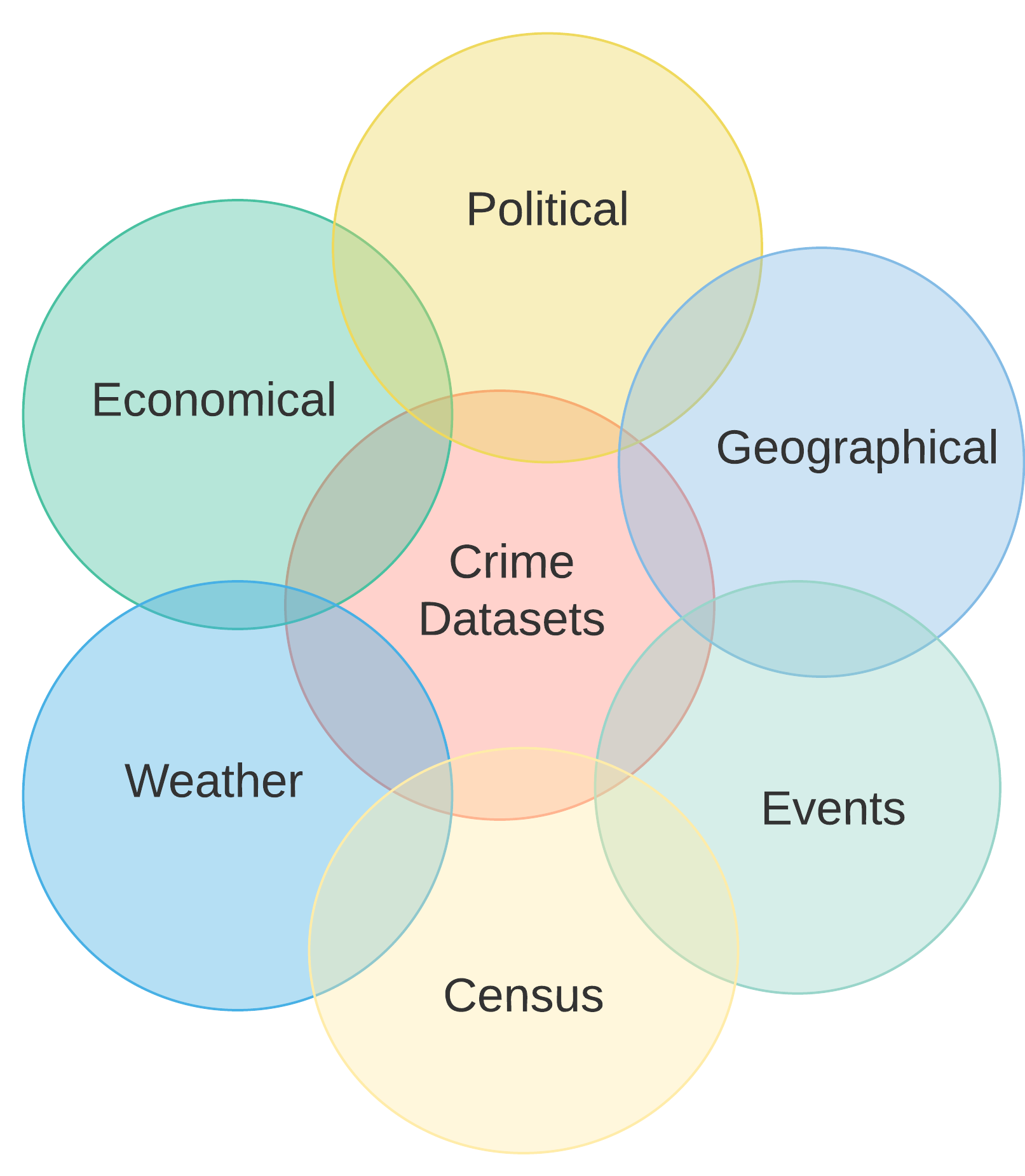}
\caption{Different sources of data needed for crime analysis}
\label{crime}
\end{figure}
Figure \ref{crime} shows data sources from different domains need to be combined with crime datasets. This combination enables analyses of crime patterns and criminal behaviors. Dealing with a variety of data from different domains requires integrating multiple data mining techniques such as classification, regression, clustering, and association rules discovery. Using multiple heterogeneous datasets or utilising multiple data mining algorithms is known as combined mining \cite{RN139}. 
\cite{RN356k,RN257} proposed hybrid predictive models for air quality prediction combining different predictors, e.g., spatial, temporal, and inflection predictors. \cite{RN257} considers the sudden change in climate as an inflecting predictor.
This interdisciplinary nature is a challenging issue that contributes to the complexity of STDM. For example, bird migrations are interrelated to climate, e.g., temperature, humidity and wind; which affects the forest areas and water quantities; which may be removed and replaced by urbanisation caused by economic development.

\subsection{Spatiotemporal Region Discretization}
\begin{figure}[]
\centering
\includegraphics[width=\textwidth]{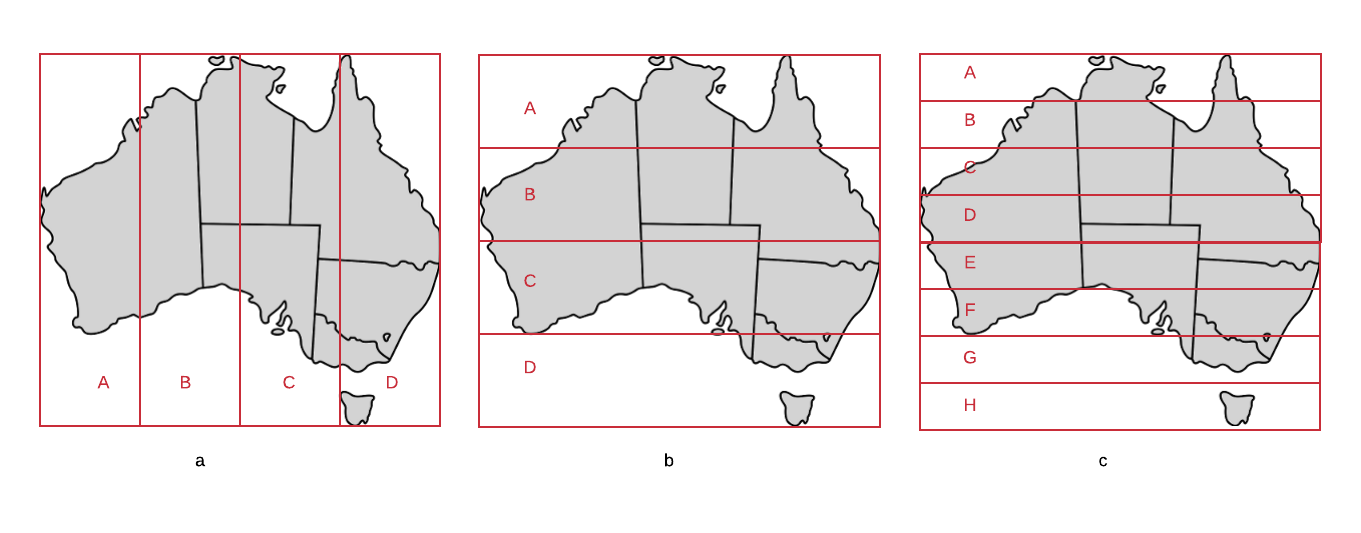}
\caption{Spatial scaling between a or b and c and zoning between a and b. The figure shows the impact of having different scales and zones on the analysis results.}
\label{MAUP}
\end{figure}

%
Before analysing spatiotemporal datasets, spatiotemporal discretisation (or aggregation) is applied. The discretisation is useful to summarise information and help in extracting features within a range rather than measuring a single point \cite{RN151}. For example, the crime rate cannot be measured for a single spatial point but requires the aggregation of the crimes occurred in wide areal units. 
Spatiotemporal patterns are scale-dependent. They shape variant clusters at different scales. For example, the crime patterns and rates are affected by the discretisation scale.
Spatiotemporal data can be aggregated at different levels or areal units. 
It is not always easy to define the best level to apply the spatiotemporal region discretisation as the results vary according to the different areal units. This is defined by \cite{RN153} as Modifiable Areal Unit Problem (MAUP). According to \cite{RN152}, MAUP includes the scale effect and the zoning effect. The scale effect would reflect the different statistical measures if the data were aggregated to different scales of areal units. The zoning effect considers the change in the borders of different areal units and their effects on the results. Figure \ref{MAUP} presents the MAUP scaling between a or b and c, and zoning between a and b. 
%
Another efficient method is to discretise the region through a regular grid with small size cells \cite{giannotti2007trajectory}. The small size is relevant to the region, i.e., the cell size could be a fraction of that region. In \cite{giannotti2007trajectory}, each trajectory is used to compute the cell densities. They calculate how many cell points intersect other neighbourhood points in the trajectory.
There are many previous works in the literature that studied the effects of spatiotemporal discretization on different applications such as remote sensing \cite{RN161}, physical geography \cite{RN162}, traffic safety \cite{RN163}, economy \cite{RN164}, health \cite{RN165}, crime \cite{flaxman2019scalable}, and ecology \cite{RN167}.

\subsection{Data Characteristics}
STDM uses spatial, temporal and non-spatiotemporal or thematic data at different levels of granularity \cite{RN139}. This fact introduces various challenging data characteristics including specificity, vagueness, dynamicity, social, networking, heterogeneity, privacy and poor quality. These characteristics are explained in the following subsections. 

\subsubsection{Specificity}
Spatiotemporal data when is used to build a good model for a particular application domain may not be useful in another one. 
%
For instance, a model that is built for the bird migrations neither be useful for vehicles movements in a city nor molecular movement in a microscopic level. This challenge is also applicable to different geographical areas having different nature and characteristics.
%
Therefore, spatiotemporal models cannot be generalised as they are designed specifically for certain domains \cite{RN66}. 

\subsubsection{Vagueness}
\begin{figure}[h!]
\centering
\includegraphics[width=\textwidth]{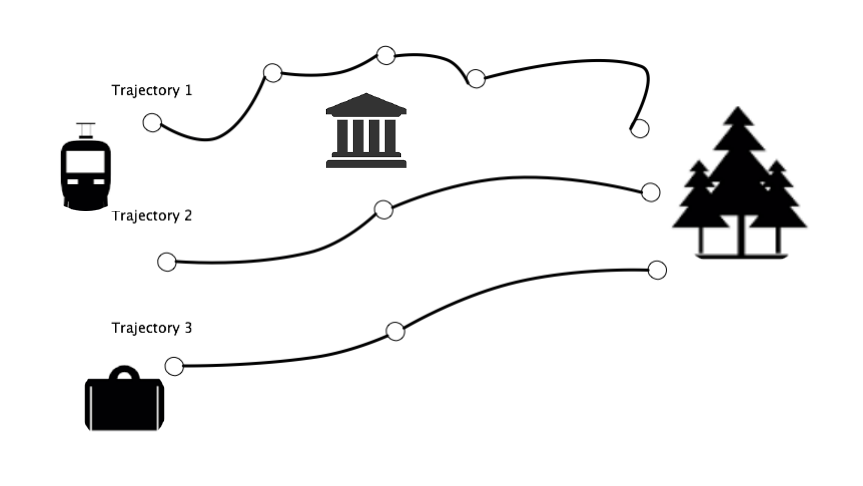}
\caption{Vagueness due to data similarities stem from different criteria. Trajectory 2 and 3 have similar spatial attributes. However, they are semantically different.}
\label{Vagueness}
\end{figure}

Spatiotemporal data objects or events have similarities that are important in different STDM tasks such as clustering. However, these similarities have different interpretations stem from different criteria. 
Two similar events may belong to different classes or be triggered by different patterns.  
Figure \ref{Vagueness} shows an example where trajectories 2 and 3 are similar from the spatial perspective, but after adding some semantic information, they become dissimilar. Because trajectory 2 departs from a train station while trajectory 3 departs from a company. In contrast, trajectory 1 and 2 appear dissimilar in terms of their spatial attributes while both of them start, pass by and end at the same location. This vagueness increases the analysis difficulty and adds further modelling and processing complexity in multiple STDM tasks such as classification, clustering and pattern extraction \cite{Shekhar2015perspective}. Therefore, there is an increasing need for more research efforts in STDM semantic annotation and enrichment. 

\subsubsection{Dynamicity}
\begin{figure}[]
\centering
\includegraphics[width=\textwidth]{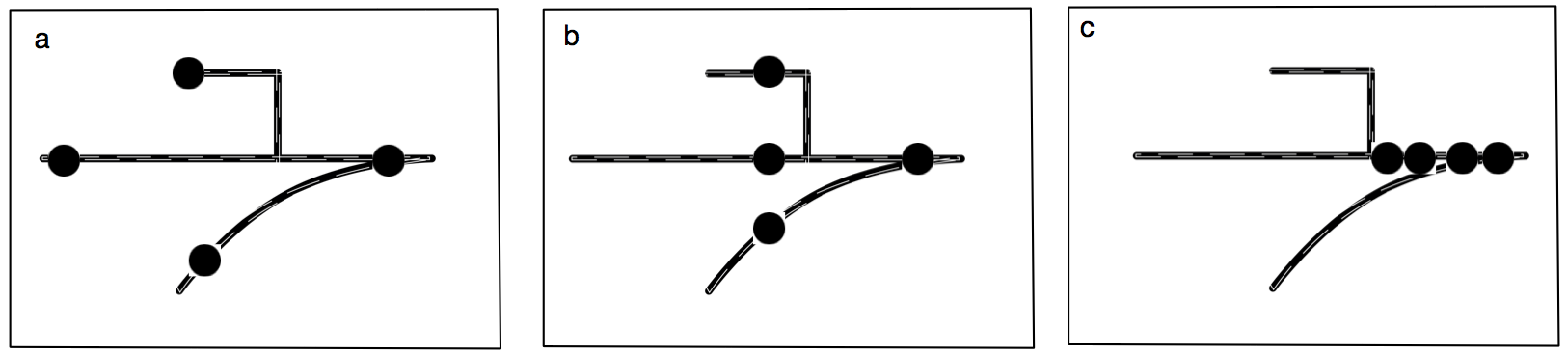}
\caption{Dynamic changing of the spatiotemporal distribution of moving objects}
\label{Dynamicity}
\end{figure}

Spatiotemporal data require dynamical models to capture the evolution of their distributions or densities. Figure \ref{Dynamicity}, a, b and c represent three different time-stamps of moving objects. As can be seen, the distribution of spatiotemporal moving objects is changing over time. This dynamic evolution of the densities can be found in different applications. \cite{RN121} described the dynamics of diseases in populations, \cite{RN69} discussed the spatiotemporal dynamics of criminal events, and \cite{RN178} raised the need to study the spatiotemporal dynamics in the case of brain transcriptome to better understand the neurodevelopment in order to predict brain disorders. 

\subsubsection{Social}
Spatiotemporal social datasets describe societies and people daily lives in different places and eras. Social media platforms contain big data related to human behaviour, traditions and people lifestyles. The social data such as text posts or tweets, images and videos are correlated with the socioeconomic characteristics. Besides, the growth of sensor technologies produces large spatiotemporal data such as check-in and geo-temporal tags. Therefore, spatiotemporal social datasets can be utilised to recognise spatiotemporal patterns in social media data or to predict social trends. Such data can also be used to discover the causes behind new social phenomena. The spatiotemporal analysis of social data is an evolving area. It has different tasks such as density estimation \cite{RN187}, collaborative filtering for recommender systems \cite{qi2020spatial,RN191} and sentiment analysis \cite{shah2019modeling}.
{
However, recent studies have shown that such models may have biasness and discrimination against different races and genders. Spatiotemporal datasets tend to have bias that affects these models. Buolamwini and Gebru \cite{buolamwini2018gender} evaluated bias present in automated facial analysis systems and datasets. They found that the analysed datasets are biased toward lighter-skinned subjects. Specifically, they categorised the gender and skin type of two facial benchmarks, namely Adience and IJB-A, according to their skin-type classification with representations of 86.2\% and 79.6\% respectivly. They also showed that darker-skinned women are most mis-classified class with 34.7\% error rate.
There is a need to capture balanced datasets that lead to unbiased systems. Bias was also investigated in employee assessment and hiring algorithms. The work in \cite{raghavan2020mitigating} studied bias in hiring systems. Specifically, they considered bias in data collection and target predictions processes. 
The study in \cite{gebru2018datasheets} proposed datasheets for datasets. These datasheets are designed to avoid bias in data collection and usages. 
}

\subsubsection{Networked}
%
Spatiotemporal data may be captured from moving objects or devices that are connected in space and time, such as GPS-tagged fleet of vehicles or animals. They form different types of networks such as in transportation \cite{RN116,han2015road}, cellular \cite{krishnan2017spatio}, wireless sensors \cite{RN124}, and smart cities \cite{RN194}. Matching the raw trajectory data with road networks makes it easier to mine the trajectory patterns. In such a case, the trajectory mining problem is done sequentially of the sequences the road network edges and stops. However, dealing with networked trajectory data is difficult due to the influence between the data points and trajectories in the network, in addition to the enormous volume of data and the exponential number of expected relationships.

When raw trajectory data, e.g., GPS readings, are matched to the road network, dealing with the resulting data is not always difficult. In the literature, such representation is considered as 1.5-dimensional and finding patterns and clustering this data is more straightforward in some contexts. For example, the problem of trajectory pattern mining could be reduced to the problem of sequential string pattern mining of the sequences of edge ids of the streets.

\subsubsection{Heterogeneity and Non-stationary}
Any environment is often affected by continuous change through space and time. Spatiotemporal data show variation in measurements and relationships due to the influences of this continuous change. For example, trajectories and behaviour of road users of a city can vary over space and time. Hence, trajectories vary for cold days compared to sunny days. This variation is known as spatial heterogeneity and temporal non-stationarity \cite{Shekhar2015perspective}.
Thus, most of the spatiotemporal data tend to have an intrinsic degree of uniqueness that may cause inconsistencies between a global model and regional models. 
Consequently, this heterogeneity requires building different mining models for varying spatiotemporal regions. Otherwise, a global model built from a spatiotemporal dataset may not describe well the observed data for a specific space and particular time \cite{RN171}. Therefore, finding the best parameters to build local models is a crucial challenge. The heterogeneity is a challenging issue that must be considered when analysing spatiotemporal data in many application domains — for instance, heterogeneity of socioeconomic observations across regions throughout social network analysis \cite{RN199}.

\subsubsection{Limited Access and Privacy}
Mining spatiotemporal data is often restricted by limited access due to privacy issues. Spatiotemporal datasets may contain information about public behaviours and norms. Service providers can mine these personal data and discover patterns and trends \cite{RN151}, which may reveal sensitive information. For example, spatiotemporal trajectories include important data about people movements, vehicles and mobile calls. 
Therefore, there is a concern with the side effects of STDM on privacy. The research on privacy-preserving STDM focuses on individuals' and personal data privacy and corporate privacy for governments and organisations.
There are different approaches to protect the privacy of the data, such as suppression of the identities of individuals, perturbation through adding noise or randomising the original data, data sanitisation, i.e., adding fake records. These methods aim to swap, modify or delete some aspect to protect the data \cite{lin2020reversible,lin2016efficient}.  In this regard, researchers face a double-edged issue, i.e., to protect privacy vs. achieving accurate analysis. 

\subsubsection{Poor Quality}
The quality of spatiotemporal data directly affects the results of the analyses. Consequently, it is important to ensure high-quality data before analysing it. This data quality assurance may not be easy to achieve when utilising interdisciplinary data that may be fragmented and distorted in disordered environments. Causes of such poor quality are uncertainties, partial knowledge, and conjectures \cite{RN139}. 
For example, STDM on bird migrations, at all times and for all locations, requires climate, water, forests data that are uncertain, sparse and reflect non-measurable aspects. 
Monitoring the physical world is affected by errors and noise that may be caused by faulty or obstructed sensors \cite{zhang2010outlier}. These errors and noise are to be corrected in order not to affect the STDM tasks \cite{tan2006introduction}.
%
\subsubsection{Big Data and Cloud Computing}
Spatiotemporal data always exist in large volumes. These large volumes are being generated by 3.8 billion people and 8.06 billion devices which are connected to the Internet \cite{Khan2018big}. \cite{villars2011big} reported 1ZB of data was created in 2010 and rose to 7ZB in 2014. This fast generation of large spatiotemporal data creates various challenges to overcome including volume, variety, and velocity. Spatiotemporal big data volume refers to the huge size that causes significant challenges in terms of storage and processing \cite{elgendy2014big}. The data volume is growing faster than the computational processing systems \cite{chen2014data}. In terms of velocity, spatiotemporal data in most applications are continuous streams of data. As such, they require expensive computational cost for processing \cite{salehian2016comparison}. These challenging characteristics of big data cause multiple issues to various STDM and applications. \cite{shao2016clustering} tackled the issue of clustering big spatiotemporal interval data, e.g., large parking data. They evaluated spatiotemporal-intervals clusters based on the similarity and balance between them. \cite{rahaman2018wait} tackled the heterogeneity issue at a large number of contextual features. They proposed a model for predicting taxi-driver wait-time at airports. \cite{shao2019OnlineAirTrajClus} utilised a big GPS data of aircraft at the airport. They proposed a framework to cluster the aircraft trajectories incrementally based on such data. \cite{ren2018location} developed a location query browse method utilising a large WiFi data of indoor physical and web activities. Such large and varied data have a spatiotemporal dependency and contextual influence on people's information and physical behaviour. 

Cloud computing has emerged to provide support to STDM to tackle different challenges in data management, storage, processing, analytics, and visualisation. Cloud computing offers large amounts of resources that enable fast, and accurate STDM. STDM computational problems are solved using new cloud technologies such as Hadoop, MapReduce, and Spark, on distributed storage systems. However, distributed frameworks suffer from many limitations in terms of data sharing, processing scalability, and interactive performance \cite{li2020big}. Besides, existing cloud solutions have limited support to the visualisation of the GIS big data \cite{wang2018visual}.
{Specifically, conventional cloud computing techniques are not designed to handle the spatiotemporal data. Therefore, spacial cloud computing techniques have recently been proposed to leverage a layer of data-as-a-service (DaaS) to virtualise the spatiotemporal data \cite{yang2011spatial}. 
The advancements of the GIS harnessed the wide availability of the cloud-based spatiotemporal services \cite{qingquan2014big}. Spatiotemporal data storage and parallel processing are widely provided by open-source cloud systems \cite{yao2018landqv2}. Other technologies such as Esri Geospatial Cloud and Google Earth Engine are also providing significant Earth observations spatiotemporal data. Overall, spatial cloud computing aims to solve spatiotemporal issues related to storage and processing. It also tends to offer better spatiotemporal data utilisation through the "as-a-service" paradigm. However, there is a difficulty in representing physical spatiotemporal phenomena that are continuous via the classical discrete based cloud approaches. This issue stems from the heterogeneity, dynamic scalability, and complex distributions of the spatiotemporal data.
}

The issues of big data and cloud computing are discussed in different surveys in the literature such as big spatiotemporal data analytics \cite{yang2019big}, social media big data analytics \cite{ghani2019social}, deep learning for big data \cite{zhang2018survey}, and big environment data \cite{sun2019can}.

\subsection{Open problems in STDM Research}\label{limitations}
There exists a wealth of research in data mining, most of which, however, focus on extracting knowledge from non-spatiotemporal data. Applying classical data mining techniques on spatiotemporal data often produces poor results \cite{Shekhar2015perspective,RN266}. Classical data mining focuses on groups of items that satisfy some rules, e.g., if events are happening together. STDM often analyses events ordered by one or more dimensions and focuses on the discovery of relationships between these ordered events, which adds more complexity to the spatiotemporal analysis. STDM deserves further research efforts to address the identified challenges and to improve the analysis methods and tools. In particular, STDM requires efforts to develop advanced data representations, modelling, visualisation, comprehensive STDM approaches, and Fairness, Accountability, Transparency, and Ethics (FATE).

\subsubsection{Data Representations}
Spatiotemporal data representation is an open research problem \cite{RN245}. There are multiple well-established spatial methods for both object-based and field-based data representation, including vector, raster data structures, spatial joins and indexing, and topological operators. These spatial data representation methods are supported in most of existing Geographical Information Systems (GIS) and spatial database systems. 
Their focus has been on the evolution of objects and fields over time in regards to discrete changes in object evolution and movement. However, this is not enough to cover spatiotemporal events and relationships \cite{Santos2016representation}. The need for new methods to represent the spatiotemporal events and relationships is important due to its impact on STDM modelling \cite{dunkel2019conceptual}.
{
Recent advances in deep learning have introduced spatially-structured networks such as graph convolutional networks \cite{hamdi2020flexgrid2vec} and recurrent neural networks \cite{rahaman2020s2rms}.
}

\subsubsection{Advanced Modelling}
One important research direction in STDM is to develop new techniques for modelling the spatiotemporal data. For instance, noticeably, most existing hotspot detection methods produce poor results when they depend only on high-density locations while ignoring the temporally related attributes, e.g., occurrence date and time, of the clustered objects. Neglecting the temporal aspects when analysing and building models from spatiotemporal data and focusing only on spatial attributes leads to unfavourable outcomes. Despite the large volume of work in GIS, there is a little support of temporal data mining in popular GIS \cite{RN205}. The temporal pattern discovery remains an under-explored area \cite{RN279}. Spatiotemporal pattern extraction methods are not able to accurately predict a pattern that may happen in a specific time because of disregarding the location and time-stamp together \cite{RN138}. 
\cite{RN291} proposed a hybrid model combining a Bayesian network and a neural network to predict car speeds. {The spatial and temporal correlation among the traffic variables led to better results.} Moreover, the thematic attributes are important to discover hidden knowledge in the spatiotemporal data. \cite{RN287} improved accuracy of visual clustering by considering both the temporal and thematic information with spatial information in their spatiotemporal data.
{Besides, modelling real-time spatiotemporal data is challenging when they are sporadically observed. This issue means that the spatiotemporal sampling is irregular such as in clinical patient time-series data. The work in \cite{NEURIPS2019_455cb265} proposed a continuous-time Gated Recurrent Unit based on the Neural Ordinary Differential Equations \cite{NEURIPS2018_69386f6b} and a Bayesian update network. The proposed methodology encodes the continuity and dynamics of the sporadic multidimensional observations. 
Capturing both global and local patterns is an essential objective of STDM modelling. DeepGLO \cite{NIPS2019_8730} is proposed as a deep forecasting method to think globally and act locally. It combines a global matrix factorisation with local temporal features.
Also, temporal information is useful in flow-estimation for applications such as video restoration. Existing methods mostly fails to capture long-range temporal features. Establishing spatiotemporal dependencies is challenging as well. Spatiotemporal Transformer Network process multiple frames at once to solve the occlusion issues in estimating the optical flow \cite{kim2018spatio}.
Generative Adversarial Networks (GAN) have also utilised for spatiotemporal modelling, simulation and data generation \cite{gao2020generative}. GANs are used to process different data types such as trajectories \cite{liu2020col,wang2020multi}, events \cite{li2019mad,yu2020extracting}, time series \cite{zhao2020adversarial,golany2020simgans} and spatiotemporal graphs \cite{gao2019progan}.
}

\subsubsection{Visualisation}
{Visualisations in data analysis are important for the decision-making as they visually summarise and present results in digestible and easy to understand forms.} Thus, it is useful to develop new approaches tailored to visualising the dynamic spatiotemporal data and the analyses results. GIS applications and current research work still focus on developing techniques for spatial visualisation, while less consideration is given to spatiotemporal \cite{RN209}. This gap requires more efforts to develop effective methods that can produce realistic, smoothing and dynamic visualisation.
{Recently, online news sources have produced large amounts of text data. One major issue in such data is visualising spatio-textual, and spatiotemporal online news trends \cite{kastner2020visualizing}. STDM methods can also be employed to visualise 3D active motions. The work in \cite{sakaue2020active} proposed to project high-frequency patterns on moving objects to visualise their 3D motion. Inspired by the human visual system, their method integrates light rays over time. Besides, spatiotemporal visualisation methods are essential to observe and analyse urban activities and behaviours. The authors in \cite{rizwan2020visualization} visualised the spatiotemporal and directional trends in urban activities. They examined both city and district levels using location-based social data. The work in \cite{salcedo2020novel} Spatiotemporal geo-visualisation method for dynamic data of the criminal activity. Data-driven approaches have recently been developed to estimate and visualise deficiencies in medical resources during the COVID-19 pandemic \cite{sha2020spatiotemporal}. 
}

\subsubsection{Comprehensive Approaches}
The nature of STDM necessitates the development of comprehensive and integrated spatiotemporal models. For example, to detect spatiotemporal crime hotspots, some other spatiotemporal tasks may be needed, such as clustering and outliers detection. Existing STDM approaches often focus on certain problems, and they do not introduce comprehensive spatiotemporal solutions \cite{RN280}. Future work in STDM ought to consider the interplay among different data types and various domains.

{
\subsubsection{Fairness, Accountability, Transparency, and Ethics (FATE)}
The attention around societal concerns of fairness, accountability, transparency, and ethics in machine learning and data mining has seen a noticeable increase recently. These concerns include amplifying genders, denying people services, and racial biases \cite{dudk2020assessing}. Web search engines results might be biased or offensive when, for example, they contain misbeliefs or posting undesirable behaviours \cite{olteanu2020when}. Retrieving the right information that are considered fair in both spatial and temporal dimensions is not an easy task as what is considered an offensive to a specific group may change over location and time.    
Buolamwini and Gebru \cite{buolamwini2018gender} evaluated bias present in automated facial analysis systems and datasets. They found that the analysed datasets are biased toward lighter-skinned subjects. The work in \cite{raghavan2020mitigating} reported that bias exists in employee-hiring systems.
The work in \cite{blodgett2020language} studied bias in Natural Language Processing systems, and found that existing methods are inferior when it comes to mitigating bias. 
People with disabilities (PWD) such as hearing impairments are directly impacted by automated systems like speech recognition systems. \cite{guo2019toward} states that such systems "may not work properly for PWD, or worse, may actively discriminate against them." 
Microsoft has released Fairlearn \footnote{https://fairlearn.github.io/} \cite{bird2020fairlearn}, a toolkit to help data scientists and developers mitigate fairness-related issues.
Later in this paper, we discuss different FATE related studies in multiple sections such as in STDM predictive modelling, public safety, mobility, environment, and Smart IoT applications. 
}

\section{STDM Task-related Challenges}\label{tasks}
STDM has different tasks, such as prediction, clustering, hotspot detection, pattern discovery, outlier analysis, visualisation, and visual analytics. 
%
These tasks are important in different applications such as understanding the behaviour of moving objects like people, birds, animals and vehicles \cite{RN233}.
%
The next sub-sections explain the challenging issues related to these STDM tasks.

\subsection{Spatiotemporal Prediction}
\begin{figure}[]
\centering
\includegraphics[width=0.99\textwidth, angle=0]{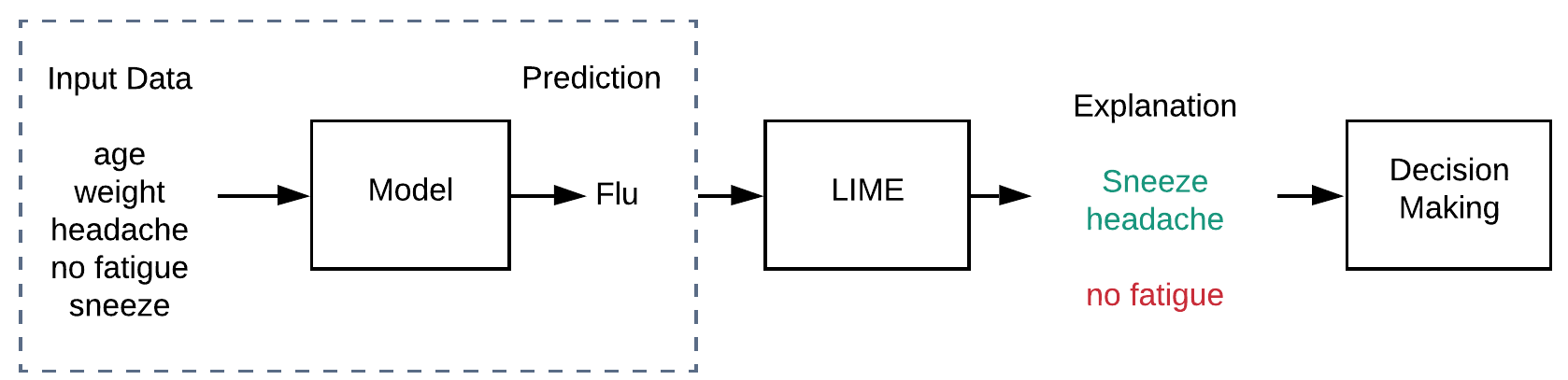}
\caption{{An example of explaining a model prediction of flue based on different symptoms, from LIME \cite{ribeiro2016should}.}}
\label{LIME}
\end{figure}

%
Data mining predictive models aim to predict target variables based on learning from annotated features of observations. These models can be either classification models or regression models. Classification models are for categorical or discrete targets, and regression models are for continuous targets. In STDM, the prediction task formulations are based on the input and output of spatiotemporal data representations. For examples, predicting an output variable, continuous or categorical, using time series at different locations in raster \cite{jia2017incremental}, predicting a scalar output using the complete information in raster data \cite{yu2015accelerated}, or predicting spatiotemporal responses using observations collected at other time-stamps in spatial neighbourhoods \cite{khandelwal2017approach}.
%
STDM prediction methods use extracted discriminative features, e.g., average speed, acceleration, duration, distance, length and direction, from labelled spatiotemporal data to train standard classifiers or regressors. The prediction can be done by single models e.g., Decision Trees (DT) \cite{Kim2015Decision}, Support Vector Machines (SVMs)  \cite{Aasha2016Effective}, or ensembles, e.g., Random Forest (RF) \cite{Phan2015Random} or deep learning, e.g., Convolutional Neural Network (CNN), Recurrent Neural Network (RNN) \cite{Liu2017Spatiotemporal,RN270,RN271,zhang2018predicting}. Other STDM tasks such as clustering can be applied to extract features for prediction. For instance, TraClass \cite{RN9t} applied segmentation and clustering for the region and sub-trajectory feature extraction, then trained an SVM classification model. STDM trajectory-based prediction estimates the future location or route of moving objects using different methods such as the Hidden Markov Model (HMM). 
Predictive models predict the next location of a trajectory to build more accurate decisions and deliver more precise recommendations. There are two different approaches proposed based on the moving object or other neighbours, as well as a hybrid approach that combines both \cite{RN25}. Recently, deep learning-based models have been applied to tackle various spatiotemporal prediction problems, e.g., crowd flow, car-hailing supply-demand, and traffic predictions \cite{ermagun2018spatiotemporal}. Here, the models take into account temporal instances, near and far spatial dependencies, and other influential external factors for the spatiotemporal prediction problem. \cite{sadri2018will} tackled another problem of predicting continuous trajectory, not only a single future location. Based on a user's morning trajectory, their model can predict the whole-day trajectory of the user.
%
The spatiotemporal dependencies between multiple contexts cause an imbalance problem as one spatiotemporal event can be rare or infrequent compared to other \cite{RN265}. This data imbalance problem affects the accuracy of the STDM prediction task.
%
The multi-scale effect or the spatiotemporal discretisation issue also poses another challenge as the results of the spatiotemporal classification or regression vary based on the different scales and zones. In order to train a prediction model, spatiotemporal features must be generated using data aggregation. However, the aggregation process can be challenging in building multiple relationships between spatiotemporal objects to build the feature sets.
Moreover, the process of generating spatiotemporal features is resource and time-consuming. Spatiotemporal Kriging is an important geo-statistical regression-based interpolation method with a spatiotemporal covariance matrix and variograms. Simply, it can predict the target values at unobserved locations based on observations at other locations, even with noisy data. However, Kriging suffers from the limitation of assuming the isotopic nature of the random variables \cite{Shekhar2015perspective}. {Recently, multiple deep learning-based studies focused on the spatiotemporal trajectory classification, such as using Long Short Ten Memory (LSTM) networks for sequence classification. However, these approaches fail to consider both spatial and temporal information simultaneously. For example, Time-LSTM handles trajectories' temporal information and neglects the spatial ones \cite{liu2019spatio}.}
{Explanation of the spatiotemporal predictions is another issue of concern. Most existing prediction models are mostly black boxes and, in many decision making applications such as medical diagnosis, understanding of the reasoning behind the predictions are required. The authors in \cite{ribeiro2016should} proposed LIME (Local Interpretable Model-agnostic Explainations), a method that explains model predictions by learning an interpretable model locally around the predictions. Figure \ref{LIME} shows the process of explaining model predictions by using different symptoms to predict that a patient has a flu. The proposed algorithm tend to identify which symptoms contribute to the model predictions. For example, sneeze and headache led to the flu prediction while 'no fatigue' is not relevant.
}

\subsection{Spatiotemporal Clustering and Hotspot Detection}
In contrast with classification, clustering partitions a set of spatiotemporal objects into similar groups based on their characteristics without having labelled datasets. STDM clustering methods aim to determine the cluster of a given object based on different features, including spatiotemporal ones. For example, trajectory clustering may harness the similarities using features such as route origins and destinations. There are different approaches used in STDM clustering such as partitioning, hierarchical and density-based. Many statistical models are also used such as HMM, spatiotemporal extensions of the density-based spatial clustering of data with noise (ST-DBSCAN), and time-focused clustering of trajectories of moving objects (T-OPTICS) \cite{RN173}. TraClus \cite{RN10} algorithm works on parts of trajectories in order to define similar trajectories based on visiting the same type of places. Clustering spatiotemporal data is affected by the large size of data which leads to a trade-off between the accurate clustering results and computational cost \cite{RN237}. It is also affected by noise and outlier patterns. Also, shapes and sizes of patterns add more complexity to spatiotemporal clustering. Spatiotemporal clustering differs according to the data types, such as clustering locations based on thematic attributes over time, clustering moving objects and clustering trajectories. Since a trajectory is a sequence of time-stamped point locations of a moving entity through space, clustering moving trajectories is complex due to their continuous object movement and evolving. Thus, more efforts are needed to discover the interaction and change in the spatiotemporal trajectory movements in order to achieve more accurate clustering \cite{RN233}. Such efforts may propose modifications of existing clustering algorithms to make them more suitable for spatiotemporal data \cite{RN109}. 
Another open issue related to spatiotemporal clustering approaches is related to their evaluation techniques. While traditional clustering approaches require computations in single Euclidean space, the spatiotemporal clustering approaches need computations in multiple spaces \cite{shao2016clustering}. Besides, computing the trajectory similarity based on point matching results in low-accuracy results \cite{li2018deep}. Specifically, such methods handle two different point-sequences in a different way albeit they belong to the same trajectory. 

%
On the other hand, spatiotemporal hotspots refer to locations that contain an unexpectedly high number of objects in a time \cite{RN294}. The spatiotemporal hotspot detection is complex because the number and features, e.g., size, shape and number of objects, of hotspots are unknown. STDM hotspot detection task is utilised for identifying dense conglomeration of events both in space and time in applications such as the outbreaks of diseases \cite{bulstra2018visceral,feng2015streamcube}. \cite{Kulldorff97Communications} proposed a spatial scan statistical (SSS) method for hotspot detection. The method explores the potential region of multiple circular-shaped sizes. The hotspot is defined as the region with a significantly high incidence of points. Later, multiple methods were proposed to generalise this SSS for spatiotemporal data \cite{Cheng2014Event}. However, the problem is still complex as results of shapes of regions, background distributions, and speeds of search. STDM hotspot detection is also utilised in various applications such as public emotion analysis, public safety, and traffic management. \cite{Zhu2016Spatio} proposed a hotspot detection method for the analysis of public sentiment using geotagged photos. The proposed method detects the emerging concentrations of certain sentiment class. \cite{Mack2018Violence} proposed a hotspot detection for political violence. The proposed method tries to solve the problem of uncertain and less predictable violence against civilians. STDM hotspot detection is employed to explore the potential locations and times of traffic accidents. \cite{Romano2017Visualizing} proposed a spatiotemporal network kernel density estimation. Using the STDM hotspot detection methods such as scan statistics and kernel density estimation are not useful in such traffic accidents cases. This is because these methods focus on Euclidean space and ignore traffic-related aspects such as constrained road networks \cite{Romano2017Visualizing}. Research studies were proposed to detect hotspots in the network space such as linear route detection which focuses on the spatial aspect and neglects the temporal dimension.
%
\subsection{Spatiotemporal Pattern Mining}
%
Spatiotemporal patterns represent the details of frequent behaviors in space and time.
STDM pattern mining works on discovering hidden information, i.e., occurrences in space and time, such as movement patterns from trajectories of spatiotemporal objects. Multiple methods were proposed to mine several types of movement patterns including; periodic or repetitive patterns that concern regular movements which are repeated at certain time intervals such as bird migration \cite{RN276}, and frequent pattern mining to discover the sequence of visited locations and the transition times between them \cite{RN277}. Trajectory pattern (T-pattern) is an example of frequent pattern defined by \cite{giannotti2007trajectory} as a set of trajectories that visit the same sequence of places consuming similar transition time. T-pattern can be mined by analysing sequences of regions of interest with time-stamps \cite{beernaerts2020spatial} or discretisation of space to determine the regions of interest \cite{giannotti2007trajectory}. Group pattern mining tends to identify movement patterns for groups of objects that move together in near space and time. Several group patterns were proposed based on spatiotemporal closeness constraint, group construction and members' properties. Examples of group patterns include flock \cite{RN15}, convoy \cite{yadamjav2019querying}, swarm \cite{shuai2019characterization}, leadership \cite{amornbunchornvej2019mining} and chasing \cite{RN19d}. They were also studied as mixed-drove or co-occurrences \cite{wang2019research,RN44} for time-unordered patterns, spatiotemporal cascades \cite{RN238} for partially time-ordered patterns, and spatiotemporal sequential patterns \cite{macikag2019discovery} for totally time-ordered patterns. Mining of spatiotemporal pattern mining has key challenges. One major challenge is that there are no explicit transactions in the spatiotemporal datasets. The number of possible patterns is exponential, and there is a potential for over-counting. Accordingly, these issues lead to a trade-off between the accuracy of the output and computational efficiency. There are different statistical methods used for mining the spatiotemporal co-location patterns, such as cross-K-function, spatial regression model \cite{RN240} and mean nearest neighbour distance \cite{RN242}. However, these statistical methods are computationally expensive due to the exponential number of candidate patterns. Furthermore, discovering spatiotemporal association or co-occurrences from trajectories is challenging due to temporal duration, different moving directions, and wrong locations.

\subsection{Spatiotemporal Outlier Detection}
In contrast to pattern mining, STDM outlier detection aims to find unusual patterns that do not follow the common path, using a set of whole trajectories \cite{RN20} or parts of trajectories \cite{RN21}. Like prediction, STDM outlier detection methods usually come after other data mining methods; especially clustering to discover objects that are not similar to any cluster.  STDM classification has also been used for outlier detection based on predefined features, e.g., location, speed, angle and direction, and utilising various distance measures \cite{RN22}. Spatiotemporal outlier detection aims to discover spatiotemporal objects that are discontinuous or inconsistent with their space and time neighborhoods. Discontinuity means that the thematic or non-spatiotemporal attributes of the outlier objects significantly deviate from other observations \cite{RN52}. Outlier detection depends on different statistical methods which are affected by several challenging issues related to model generalisation and scalability, lack of effective spatiotemporal data representation methods, and low focus on interpretability. Although the mathematical and statistical foundation of spatiotemporal outlier detection is important, most previous researches focus only on the computation efficiency and intuitive analysis \cite{RN174}. Additionally, spatiotemporal outliers can be important and refer to interesting events, e.g., the formation of cyclones, or they can be noise \cite{RN245}.

\begin{figure}[]
\centering
\includegraphics[width=0.99\textwidth, angle=0]{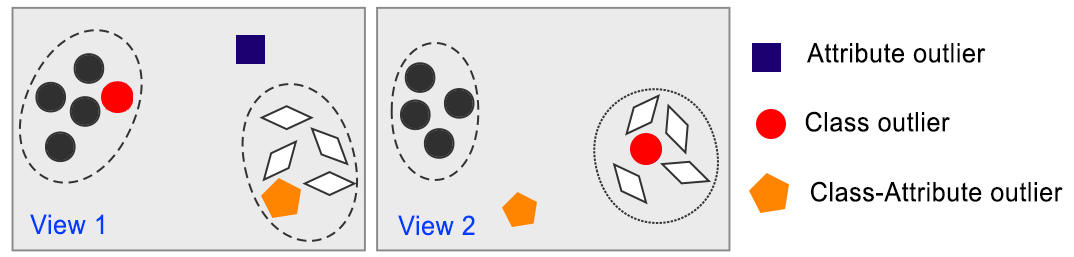}
\caption{{Three different types of outliers \cite{ji2019multi}.}}
\label{mutliViewOD}
\end{figure}

{A new research trend in outlier detection is multi-view outlier detection, i.e., a multi-view learning task. This task is challenging due to the complex distributions of data across different views. It focuses on three outlier types, including attributes, class, and class-attributes outliers. There is a need to accomplish such multi-view outlier detection because most existing approaches consider part of the problem \cite{ji2019multi}. Figure \ref{mutliViewOD} illustrates the three different types of outliers. The attribute outlier, red triangle, is an example of abnormal behaviour. The class outlier, blue circle, behaves normally in each view but abnormal across different views. The class-attribute outlier, green square, represents an attribute outlier in some views and class outlier in other views.}

\subsection{Spatiotemporal Visualisation}
Spatiotemporal visualisation task employs techniques for spatiotemporal data presentation. These techniques go beyond static or traditional 2D maps to include modern 3D spatiotemporal cubes and interaction methods to uncover the implicit spatiotemporal knowledge. Spatiotemporal visualisation is discussed within various applications such as marine environment \cite{RN210}, news events \cite{RN211}, social topics \cite{koylu2019modeling}, urban dynamics evaluation \cite{xia2020understanding,RN258} and mobile data \cite{RN214}. Despite much previous work on spatiotemporal visualisation, there are still unsolved issues such as visualising big spatiotemporal data in real-time \cite{RN248}. The challenge stems from the difficulty of temporal representation on maps because of the limitation of GIS in representing dynamic processes. This issue is further exacerbated by the fact that most geographic phenomena vary over time. There are many proposed methods based on 2D maps, such as showing small charts on maps \cite{reza2019hi}. However, this cause issues such as overcrowded maps when plotting many charts, and overlapped charts with close locations. Moreover, 3D space-time cubes also pose additional challenges. For example, it is difficult to represent space-time paths to geo-locations and time due to the 3D display, as well as the difficulty to display big spatiotemporal datasets.

\subsection{Spatiotemporal Visual Analytics}
STDM task of visual analytics considers identifying significant locations and periods along with movements such as a bus route \cite{Mazimpaka2016Visual}. STDM visual analytics methods harness the effect of significant spatial features during certain temporal periods. The selected discriminative spatiotemporal features are used to classify locations overtime periods. There are multiple applications in which knowing such locations and periods is crucial such as in traffic management and route planning. The visual analytics of public transportation has received less research focus in comparison to taxis or private cars \cite{Mazimpaka2016Visual}. STDM visual analytics methods were proposed in multiple research works such as in \cite{doraiswamy2018spatio,Lv2012Discovering,Bhattacharya2012Extracting}. However, these studies focused on spatial analysis and neglected the temporal dimension. A major challenging issue in STDM visual analytics is to visualise both spatial and temporal information at the same time. Multiple methods were proposed to solve this STDM visual analytics problem such as space-time cube \cite{Tominski2012} and multiple 2D maps \cite{Wang2014Urban}. However, most of the proposed methods focus on discovering global patterns between origin and destination. There is a need to discover local patterns at stops and segment levels.
%

{
\subsection{Computer Vision Related STDM Tasks}\label{CVtasks}
Computer vision research aims to extract useful information for images. Spatial visual feature extraction is essential in various tasks such as image classification and hand writing recognition \cite{al2018arabic,al2017enhanced}. Spatiotemporal based STDM focuses on learning how to see a scene, understand components, and track moving objects in a video. The complexity of the visual world makes STDM challenging in computer vision tasks. STDM methods has been utilised most computer vision tasks such as tracking \cite{huang2019re,wen2019learning,yin2019capturing,Bai_2019_CVPR} and segmentation \cite{Xu_2019_CVPR,xu2019mhp,Wang_2019_CVPR,Hu_2019_CVPR}. In this section, we focus on recent challenges in visual tracking and segmentation, specifically, drone-based object tracking and amodal semantic segmentation, respectively.
}

\begin{figure}[]
\centering
\includegraphics[width=0.7\textwidth, angle=0]{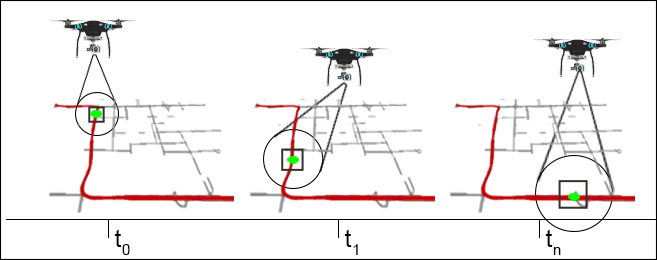}
\caption{
Drone-based object tracking \cite{Hamdi2020DroTrack}.}
\label{drone}
\end{figure}

{
Visual object tracking (VOT) is a key component of multiple domain application such as surveillance, search and rescue, and topographic mapping. VOT is a challenging task due to visual noise, occlusion, cluttered backgrounds, and dynamic variation of moving object features. VOT methods aim to track the moving objects temporally in a video and spatially over the frame pixels. There are large bodies of research in both single VOT \cite{follmann2018mvtec,Kart_2019_CVPR} and multiple VOT \cite{tang2017multiple,zhou2018deep}. New VOT tasks include object tracking with segmentation \cite{follmann2018mvtec}, tracking by reconstruction \cite{Kart_2019_CVPR}, graph convolutional tracking \cite{gao2019graph}, 
deep multi-scale spatial-temporal tracking \cite{zhang2020non}
and drone-based VOT \cite{yu2020conditional}.
In drone-based object tracking \cite{Hamdi2020DroTrack}, a drone \emph{d} is tracking a moving object in real-time \emph{o} using a camera as illustrated in Figure \ref{drone}. Unlike conventional object tracking using fixed cameras, a camera mounted on \emph{d} is moving according to the motion of \emph{d}.
When \emph{d} or \emph{o} moves the distance between them is altered. This fact leads to changes in the location and scale of \emph{o} in the video frame.
Figure \ref{drone} shows three different tracking positions of a drone in different time-stamps monitoring a moving object indicated in light green. As illustrated, the scale of the moving object is inversely related to the size of the drone’s field of view. When the drone flies high and has a wide field of view, the object becomes smaller. Conversely, the object scale is enlarged if the drone gets close.
New VOT drone-based dataset have been released such as
DTB70 \cite{li2017visual}, 
UAV123 \cite{mueller2016benchmark}, 
UAVDT-Benchmark-S \cite{du2018unmanned}, and 
VisDrone2019-SOT \cite{zhuvisdrone2018}. 
The datasets are of high diversity and captured in multiple environments. They cover more difficulties and aspects that are not found in the traditional tracking datasets such as VOT \cite{kristan2014visual} and VTB50 \cite{wu2013online}. The drone-captured datasets include both translation and rotation camera motions. 
The literature shows that these datasets are challenging for conventional tracking algorithms. They also cover highly challenging cases in both short-term and long-term occlusion. The datasets contain different moving object types, such as humans, animals, cars, boats, birds and drones. This variety offers different levels of degree of freedom for the motion. Objects like cars and boats can only translate or rotate, whereas humans and animals, birds and drones have a higher degree of freedom. The datasets outdoor scenes are in various situations, including significantly varied backgrounds. These challenging motion characteristics cause object deformation, leading to more difficult object tracking. 
}

\begin{figure}[]
\centering
\includegraphics[width=0.8\textwidth, angle=0]{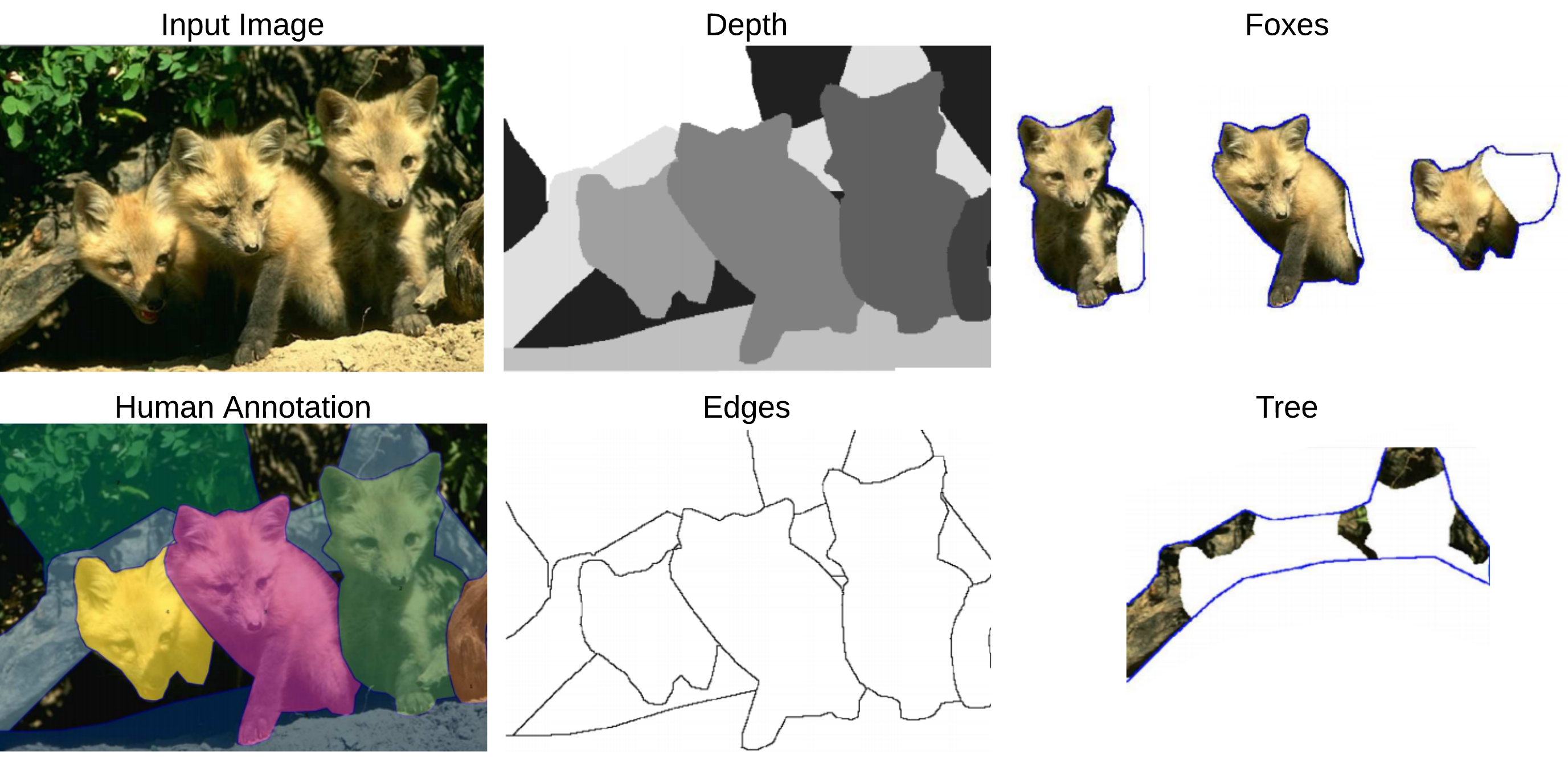}
\caption{{
An example of semantic amodal visual object segmentation \cite{zhu2017semantic}. The first row shows the original scene and its segments human-annotation. The second row visualises the depth and visible edges. Finally, the third one shows the semantic annotation of the invisible regions.}}
\label{amodal}
\end{figure}

{Video object segmentation (VOS) extracts foreground objects in a video. VOS is a fundamental task for many video analysis tasks such as video summarisation and understanding. Most of VOS related work is either unsupervised, i.e., does not require human annotation \cite{tokmakov2017learning,li2018unsupervised,hu2018unsupervised} or semi-supervised, i.e., requires to annotate object in the first frame only \cite{cheng2018fast,ci2018video}.
VOS maintain the temporal associations of object segments through the video usually using optical flow \cite{bao2018cnn,hu2018motion}. It aims to model the pixel motion over time. However, optical flow annotation requires expensive human efforts and is not always suitable for VOS. Recently, VOS researcher proposed end-to-end trained deep neural networks to overcome such issues, such as  spatiotemporal sequence-to-sequence network \cite{xu2018youtube} and deep recurrent network \cite{li2018video}. There are multiple datasets for video object segmentation such as 
DAVIS16 \cite{perazzi2016benchmark}, 
FBMS \cite{ochs2013segmentation}, 
JumpCut \cite{fan2015jumpcut}, 
Youtube-Objects \cite{prest2012learning}, 
SegtrackV1 \cite{tsai2012motion}, 
and instance segmentation data such as
Youtube-VOS \cite{xu2018youtube}, 
SegtrackV2 \cite{li2013video}, and
DAVIS17 \cite{pont20172017}.
However, none of these datasets offers direct learning of new research directions in VOS such as Semantic Amodal instance level (SAVOS) \cite{hu2019sail}. SAVOS aims to segment individual objects in a video under occlusion semantically. Figure \ref{amodal} presents an example showing the steps of segments annotation, depth and visible edge estimation, and annotating the edges of the invisible regions. SAVOS is a useful task for object size prediction, depth ordering, and occlusion reasoning. This task requires the temporal sequence, in the video dataset, to be densely and semantically labelled. Such data should be essential to analyse the object and human motion behaviours. New datasets have been released recently to fit the SAVOS task such as \cite{maire2013hierarchical,zhu2017semantic,ehsani2018segan,follmann2018mvtec}. Human is able to predict the occluded parts with confidence and consistency \cite{zhu2017semantic}. However, this task is still a challenging task in STDM.
}

\section{STDM Application-related Challenges}\label{domains}
In addition to the earlier above-mentioned spatiotemporal challenging issues, this section describes application-related issues. The following sub-sections discuss six major STDM applications including, crime and public safety, traffic and transportation, earth and environment, epidemiology and spread of infectious diseases, social media analysis and smart Internet of things (IoT).

\subsection{Crime and Public Safety}
Crime data varies and have interesting characteristics that motivated previous works. However, the domain of public safety has its challenges such as the lack of comprehensive and generalised analysis methods that can handle complex and heterogeneous data types such as historical, geographical and demo-graphical data. Furthermore, there is a lack of systematic analysis and representation of the temporal crime attributes, as well as the unavailability of systematic literature reviews \cite{RN215}. 
%
In addition, the dynamic nature of the crime patterns is affected by opportunities and the existence of motivated offenders and suitable targets \cite{RN150}.
%
Hence, many researchers focus on deriving features from human activities to tackle the problem of crime event prediction. Some studies use mobile network data to derive human dynamics and aggregate them with other factors, such as demographics, to predict crime events. Other works analyse social media, e.g., Twitter and Foursquare, data for crime prediction \cite{RN345}. However, they often utilise aggregated datasets over long periods. This challenge raises the need to develop quantitative methods that work on high-resolution data \cite{RN69}. Moreover, the new methods need to consider characteristics that are specific to spatiotemporal crime data such as the susceptibility of having outliers and noise.
{Recent literature in crime analysis utilises the spatial correlation in fine-grained crime modelling. The continuous conditional random field is used to capture the relationships among different regions. However, it can not deal with dense graphs considering the potentially large amount of nodes and relations in a graph of a fine-grained level. Deep neural networks are utilised to reduce the model complexity and improve the training accuracy \cite{yi2019neural}. 
}
{Crime prediction models try to answer the questions of where and when the next crime may occur. However, as discussed earlier, such models may be biased toward genders or races. Racial bias in predictive policing is a cumbersome issue. Multiple (non)-governmental organisations are concerned about bias in low enforcement applications and fear that predictive methods may target minority communities.  
There are studies that show the existence of the racial bias in different public safety applications such as pedestrian stops \cite{legewie2016racial}, racial profiling of vehicles \cite{horrace2016dark}, use of force \cite{ferguson2019rise}, and drug enforcement and arrests \cite{lynch2013policing}. The work in \cite{richardson2019dirty} investigated the existence of racial bias in predictive policing. The work studied the link between illegal bias in police practice and dirty data that are used to train predictive models. 
}

\subsection{Traffic and Transportation}
Unlike the public safety domain, previous works in traffic and transportation focus on disaggregated micro-scale data capturing a large number of observations. Thus, dealing with such spatiotemporal datasets adds more challenges related to the collection, storage and processing of such big and dynamic data, then using them to build comprehensive spatiotemporal models. Besides, thematic data related to traffic accidents, injuries and road networks combined with criminal records is an important multidisciplinary path of research that needs more efforts. Traffic data represents spatiotemporal trajectories that are used to discover periodic patterns that describe the behaviours of moving objects \cite{sonmez2019novel}. One important challenge is that the spatiotemporal trajectory pattern does not follow regular time intervals \cite{RN219}.
Furthermore, the influence of nearby objects and their patterns is another problem. Example of such influence is spatiotemporal events, such as accidents, that affect the traffic congestion patterns \cite{RN52}. Traffic congestion estimation is another open issue due to the complexity of analysing multiple data from different sources, e.g., sensors in taxicabs, GPS, mobile sensors, and road network sensors, and the inclusion of various variables such as density, velocity, inflow and previous status \cite{RN115}. Adding to these computational issues, traffic congestion is a critical problem because it affects peoples' life and may damage the socioeconomic growth \cite{Zheng2014Urban}.

\begin{figure}[]
\centering
\includegraphics[width=0.8\textwidth, angle=0]{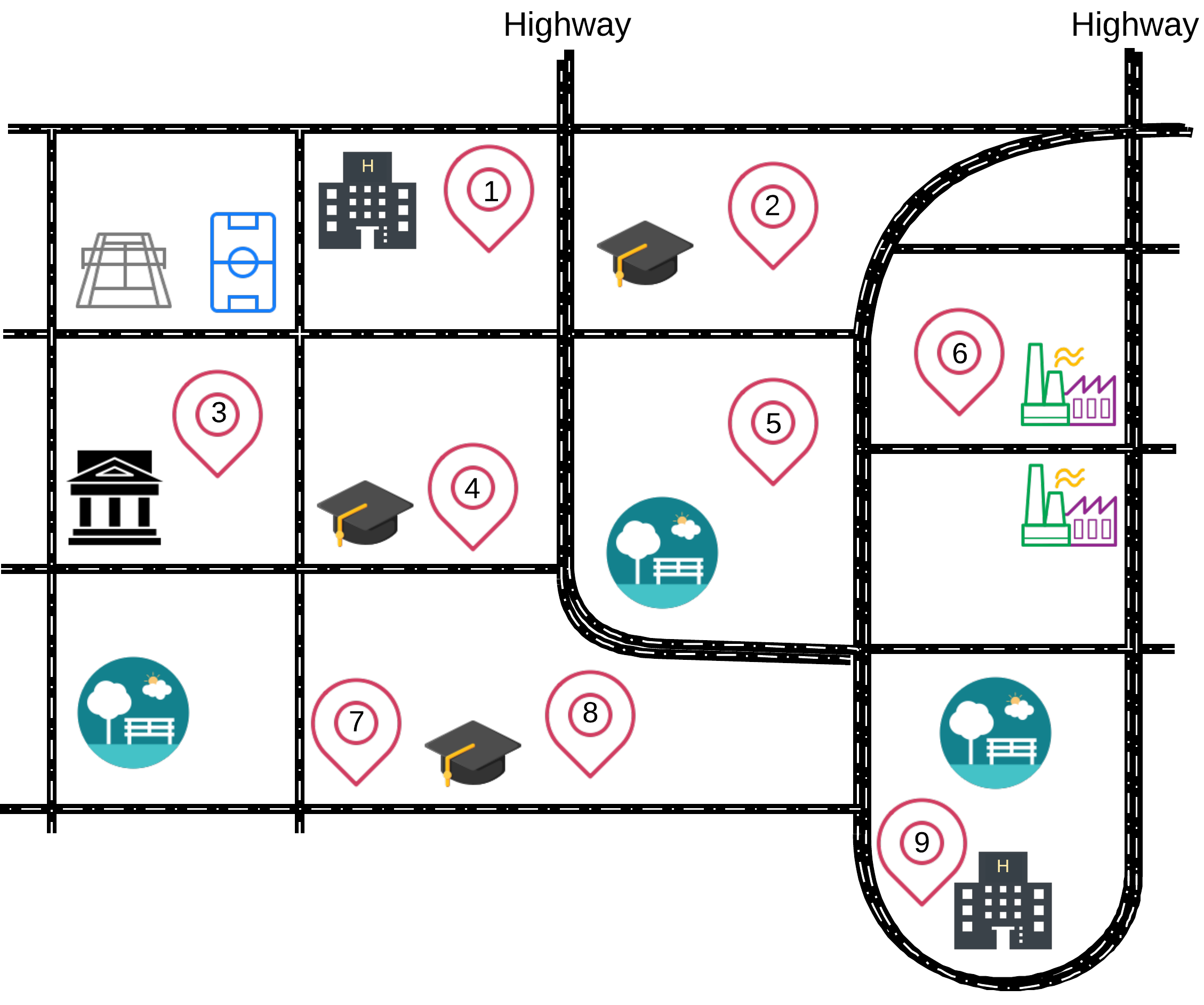}
\caption{The impact of different correlations among regions on RLRH demand forecast. For example, R7 is adjacent to R8, similar to R4 and R2, connected with R3, and distant or irrelevant to R6.}
\label{RLRH}
\end{figure}

{Deep learning researchers have paid attention to traffic state estimation due to the availability of large datasets of vehicle trajectories \cite{zhang2019boosted}. There are two main approaches to address traffic estimation. The macro approach divides the city into equal grids which are represented by its road segments states \cite{zhang2016dnn,vahedian2017forecasting,yao2018deep}. The micro approach produces finer-grain segments of road networks \cite{yu2017deep}. This approach considers temporal traffic patterns and spatial correlation among traffic states. Traffic estimation is still a challenging task due to the uncertainty issue because of data sparsity and semantic ambiguity. Trajectory embeddings are tackled by the complicated topology of the transportation networks. Modelling such data is affected by the spatiotemporal dynamicity.
Moreover, intelligent transportation systems have other essential components, such as Region-Level Ride-Hailing (RLRH) demand forecasting. RLRH demand forecasting aims to estimate the future demand in city regions given the previous states. RLRH demand forecasting tends to enhance traffic functions such as vehicle allocations, waiting time, and congestion \cite{yao2018deep}. In the same fashion, this task is challenging because of the complex nature of the STDM correlations or complicated dependencies among different regions. Figure \ref{RLRH} shows an example of the impact of the different correlation level of different regions on the RLRH demand forecast \cite{geng2019spatiotemporal}.
}
{
Moreover, recent studies focused on the fairness in the ride hailing platforms such as Uber and DiDi as they employ different matching strategies to connect customers and drivers. There seems to be an unfair distribution of jobs among drivers which led to concerns such as discrimination against minorities. Having each matching to be fair is a difficult task. Therefore, there is a need for more efforts in STDM to discover better matching distributions over time. The authors in \cite{Tom2020Fairness} proposed a framework that attempts to ensure fairness in the ride hailing matching. Their hypothesis of fairness is that all active drivers should be proportionally matched overtime. 
}

\subsection{Earth and Environment Monitoring }
There are different environment-related spatiotemporal applications such as land use and change detection that are being affected by different natural and socioeconomic factors. Another application area is route mining in waterways, which is challenging due to routes may be created by different types of ships jamming in the same waterway, frequent changing direction and navigation via different routes \cite{RN213}. There is a need to integrate data describing the atmosphere, hydrosphere, lithosphere, and biosphere of the earth in order to build more accurate spatiotemporal multidimensional models \cite{RN52}. Also, there is a lack of mathematical and statistical methods for dynamic visualisation \cite{RN209}. Quantifying spatiotemporal human exposure to air pollutants is another difficult task because of different people activity patterns and the complexity that results from multifaceted relationships between human and environment \cite{RN226}. 
{
Spatiotemporal climate forecasting predictive models are affected by errors in their model physics \cite{meehl2014decadal,hamdi2021drone}. Specifically, models drift toward their internal means state. This issue is also caused by imperfections in representing the models' initial conditions \cite{hazeleger2013multiyear}. Recently, bias correction methods have been developed. There are multiple methods that are proposed to solve this bias in spatiotemporal predictive modelling, such as bias adjustment and conditional bias methods. These methods correct spatiotemporal bias in the climate forecast time, lead time, and initial conditions \cite{director2017improved}. The significance of this STDM task extends to commercial vessel traffic in some Arctic regions. Such commercial domain is directly impacted by the sea ice forecast \cite{huntington2015vessels}.
Spatiotemporal bias also exists in precipitation data gauging and analysis. Gauge-based rainfall predictions rely on point data which are collected from multiple areas with limited and uneven radius. Satellite-based rainfall predictions have been developed using deep learning bias correction models \cite{le2020application}.
}

\subsection{Epidemiology and Spread of Infectious Diseases}
\begin{figure}[]
\centering
\includegraphics[width=0.6\textwidth, angle=0]{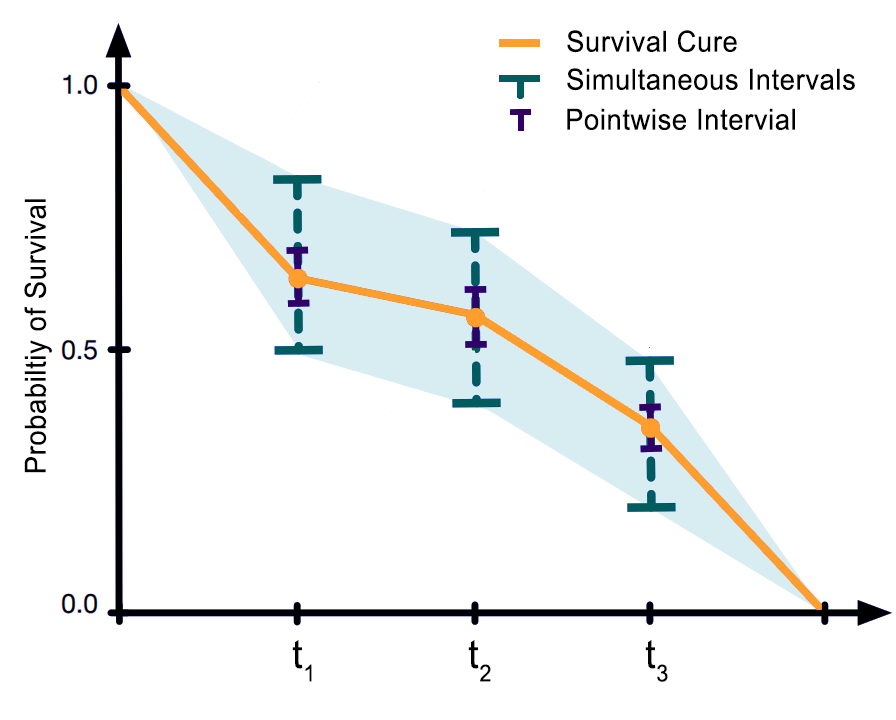}
\caption{{Survial analysis under uncertainty \cite{sokota2019simultaneous}. The survival curve, red line, calculates the probability as a temporal function. The point-wise and simultaneous intervals covers the uncertainties.}}
\label{survival}
\end{figure}

The spread of infectious diseases is affected by human mobility. Monitoring human mobility is a hard task because it occurs in huge volumes, and different periods ranging from minutes to years \cite{RN121}. Another challenge is the enormous number of mobility events that have different characteristics, causes and complex spatiotemporal relationships with humans. On the other hand, migrants and refugees who come from regions of conflicts add more challenges to investigate new factors from outside the analysed society or geographic areas. The migrants carry their health history, which may affect the population of the target societies. For example, the black death disaster came to Europe via the silk route. \cite{RN121} reported that there are two main problems in the spatiotemporal spread of pathogens of humans; namely; fast spatial spread of an evolving pathogen and the interactions of health systems. The modern transportation facilities enable this fast spread of infectious pathogens around the world. Besides, pathogens do not care about borders in contrast with the public health systems that manage a specific geographic area. Consequently, the interconnections between different areas add more challenge in mining the spread of pathogens. {For example, the outbreak of 2019 new corona-virus diseases (COVID-19) in Wuhan, China, has forced the countries all over the world to close their borders and apply strict travel bans \cite{novel2020epidemiological,pan2020time}. 
Recently, attention models with deep LSTM networks have been employed to predict disease progression \cite{zhang2019attain}. This task considers irregular time intervals between consecutive disease events. 
Moreover, understanding the probability of patient survival is a piece of essential information for the healthcare field. It is useful to identify the best treatment plans over time. Survival analyses consider the prediction of occurring an even of interest. However, these analyses are affected by spatiotemporal uncertainty. Therefore, most of the existing survival analysis methods lack the ability to provide comprehensive results. Figure \ref{survival} shows a representation of calculating the uncertainty of survival analysis in temporal function \cite{sokota2019simultaneous}.
}.

\subsection{Social Media Analysis}
Data provided by social media is affected by the growth of sensor technologies that generate big spatiotemporal data, such as check-in records, user reviews, and geo-temporal tagged posts \cite{RN219}. In this context, social media has complex spatiotemporal correlations. Accordingly, the analysis of social media faces different challenges in order to understand these correlations and to develop accurate models. Spatiotemporal topic modelling in social media contents is used with time and location-tagged to discover topics. This topic modelling is affected by challenges related to the heterogeneity of geographical context \cite{jiang2018neural}, such as the locations sparsity caused by a tiny amount of posts that are tagged with geographical locations. The sparsity issue directly affects spatiotemporal social media analysis tasks such as density estimation \cite{jeawak2020predicting,RN186}, event location extraction \cite{RN281} and collaborative filtering \cite{zhou2019spatio}. 
{
Such tasks are affected by different social factors such as social trust. Trust propagation approaches estimate the users' preferences based on the features of their connections. This could achieved by implementing random-walk on social graphs. Recently, deep learning studies have merged social information into recommendation in an ensemble fashion such as graph neural network \cite{song2019session,wu2019dual}  and network embedding \cite{liu2018social}. These works aim to encode high-dimensional network information \cite{wu2019feature}.
Most of the existing methods predict the social relationships based on pairwise approaches with hand-engineered features or utilising skip-gram model to learn the graph embeddings. However, both models fail to harness complex dynamics of social relationship patterns. Besides, computing the graph embeddings based on random walks for information propagation is not accurate due to lacking the semantic information \cite{wu2019graph}.
}

\begin{figure}[]
\centering
\includegraphics[width=0.9\textwidth, angle=0]{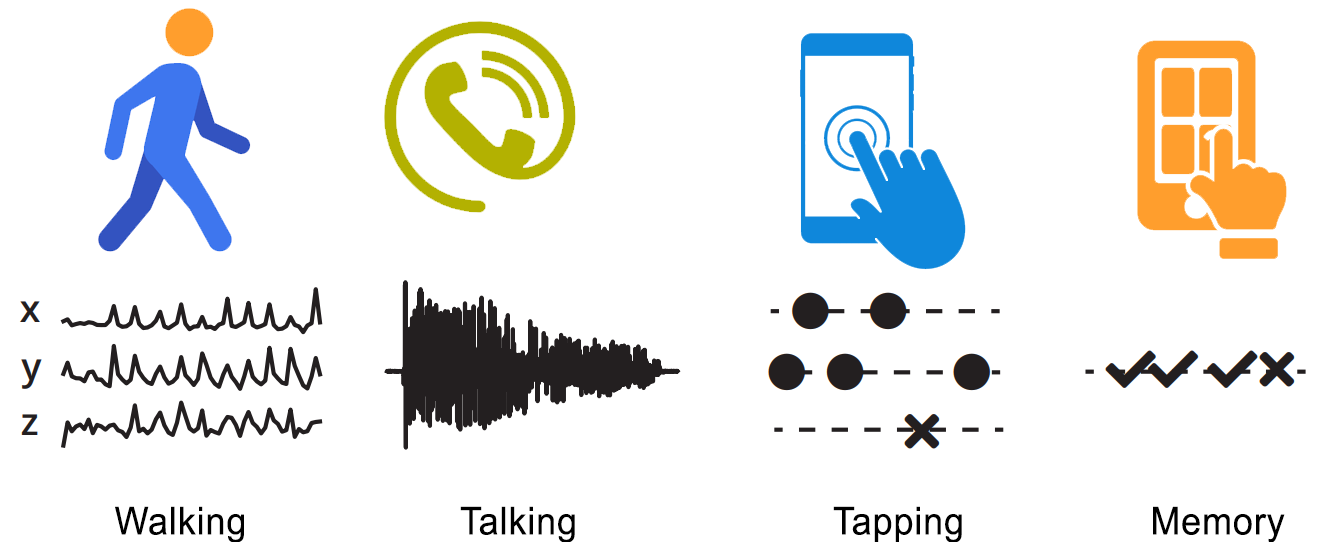}
\caption{{Mobile-collected data can be used to monitor different patterns of user's walking, voice, tapping, and memory \cite{schwab2019phonemd}.}}
\label{Parkinson}
\end{figure}

\subsection{Smart Internet of Things}
Smart IoT adoption is increasing with the proliferation of urban population. Consequently, real-time large-scale sensor data are captured via distributed connected objects equipped with sensors. These sensors can monitor different variables at multiple locations such as homes, ambient air quality, climate and earthquakes \cite{RN356k,RN79,mehrjoo2018accurate,RN81,yunus2020decadal}. Working with such data poses some challenges while storing, processing and modelling, which can be addressed by utilising cloud computing \cite{RN2}. Sensory data suffers from sparsity where no observations are captured for many regions within the sensed fields. This requires more efforts in filling the missing values using correlated data sources. However, the integration process is complicated because of the different sampling rates of various sensory data. For example, the temporal resolution in medical sensors is much higher than in GPS sensors. Also, the sensory data integration is complicated by the veracity of the different measurements recorded by each sensor. For instance, there is a high level of uncertainty caused by the difficulty in spatial localisation and temporal synchronisation between sensors \cite{RN82}. 
{
IoT opens new research directions, such as sitting posture monitoring and smart voice control. The poor sitting posture causes ill-effects on both physical and mental health. Sitting posture monitoring is challenged by multiple issues such as the variance in human sitting behaviours and divergence in user body mass indexes \cite{bourahmoune2019ai}. Smart voice control is being used by 100 million users \cite{lei2019design}. Different users can use multiple appliances at the same time. There is a need for new STDM methods that can consider the possible variant spatial, temporal, and thematic features. For example, voice-controlling a smart air-conditioner may be different from a smart TV. It is expected to have different user behaviours based temperature degrees and device locations. 
Moreover, IoT mobile devices can be used as a smart agent to detect certain diseases such as Parkinson's disorder. This disorder cause degeneration of the nervous system. It affects human movement, speech, and cognition. Clinical assessments are commonly misdiagnosed such disease. However, this could be solved by analysing long-term data that can be collected using mobile devices \cite{schwab2019phonemd}. Figure \ref{Parkinson} shows an example of Parkinson detection based on mobile-collected data of walking, talking, tapping, and memory. 
}
{
Another issue in multi-sensory data fusion is to eliminate spatiotemporal bias. Fusing data from multiple sensors suffers from time delays among measurement timestamps. Such delays might be caused by data transfer or signal processing. Therefore, data fusion applications are in need for accurate spatiotemporal bias estimation techniques \cite{bu2019simultaneous}.
Various methods are proposed for different sensors such as cameras, radars, or sonars \cite{taghavi2016multisensor,jones2011visual}. They utilised different computing methods such as maximum likelihood and Kalman filters. However, these works only consider spatial data references of the multiple sensors. The temporal bias inevitably exits in the multi-sensor data fusion and requires more research efforts \cite{bu2019simultaneous}.
}

\section{Summary of STDM General Challenges}\label{summary}
Table \ref{table1} summarizes the identified and above-mentioned STDM challenges. For each challenge, the table highlights its causes and related works attempting to address them.  
Furthermore, the related works are annotated with relevant STDM tasks and applications. The aim of this annotation is to link between the three main components of the survey. For instance, STDM relationships pose three main challenges including complexity, implicitness, and non independent and identically distributions. Each of these challenges is accompanied with its tasks and applications. \emph{Tasks} include Cluster Analysis (CA), Pattern Analysis (PA), Outlier Detection (OD), Prediction Modelling (PM), Visualization (v), and Visual Analytics (VA). Also, the \emph{Applications} are Public Safety (PS), Transportation Management (TM), Environmental Analysis (EA), Epidemiology (EP), Social Media Analysis (SA), and IoT.
\afterpage{
\begin{longtable}{|p{1.5cm}|p{1.75cm}|p{2.75cm}|p{1.75cm}|p{1.75cm}|}
\caption{Summary of STDM main challenges and their causes}
\label{table1}
\setlength{\tabcolsep}{0.5em} 
\\ \hline
Issue & Challenges & Causes & Tasks & Applications\\ \hline
\multirow{5}{*}{\noindent\begin{tabular}{p{1.5cm}}Spatio-temporal Relation-ships\end{tabular}} & \multirow{2}{*}{Complexity} & 
Discrete representation of continuous spatiotemporal data &S-\cite{RN52,Shekhar2015perspective}, CA-\cite{RN287,RN233}, PA-\cite{cheng2020high}, OD-\cite{RN288}, PM-\cite{liang2019deep,zhang2019flow,RN291}, V-\cite{RN293}& PS-\cite{RN256}, TM-\cite{RN345}, EA-\cite{RN350}, EP-\cite{RN337}, SA-\cite{RN347}, IoT-\cite{RN345} \\ \cline{3-3}
 &  & Co-located objects influence each other &   &   \\ \cline{2-5} 
 & Implicit-ness & Implicit relationships between spatiotemporal objects   & CA-\cite{RN307,kang2020urban}, PA-\cite{RN340,RN306}, OD-\cite{RN308}, PM-\cite{RN281} & PS-\cite{RN349}, TM-\cite{RN342}, EA-\cite{RN350}, EP-\cite{RN339}, SA-\cite{RN340}, IoT-\cite{RN340} \\\cline{2-5}
 
 & \multirow{2}{*}{\begin{tabular}{p{1.2cm}}Non-independ-ent and Non-identical distrib-ution \end{tabular}} & 
 Auto-correlation due to dependency relationships in space and time & S-\cite{Shekhar2015perspective}, CA-\cite{RN313,guijo2020time}, PA-\cite{shirowzhan2020data}, OD-\cite{RN313}, PM-\cite{he2019stann,RN237}, V-\cite{RN315} & PS-\cite{RN326}, TM-\cite{RN341}, EA-\cite{RN351}, EP-\cite{RN338}, SA-\cite{RN352}, IoT-\cite{RN335} \\ \cline{3-3} 
 &  & Non-identical distribution across space and time &  &  \\ \hline
 
\multirow{3}{*}{\noindent\begin{tabular}{p{1.5cm}} Inter-disciplin-ary and Combined Data Mining \end{tabular}} &
Various interrelated domains &
\multirow{3}{*}{\noindent\begin{tabular}{p{2.7cm}}Heterogeneous data requiring multiple STDM techniques \end{tabular}} & CA-\cite{RN318,shao2019OnlineAirTrajClus}, PA-\cite{RN317}, OD-\cite{RN319}, PM-\cite{RN237}, V-\cite{RN316} VA-\cite{malik2014proactive}, & PS-\cite{RN325}, TM-\cite{RN341,RN317}, EA-\cite{RN351}, EP-\cite{RN331}, SA-\cite{RN334}, IoT-\cite{RN353} \\ \cline{2-2} 
 & Environ-mental factors &  &  &  \\ \cline{2-2}  
 & Opportunity &  &   & \\ \hline
 
\multirow{2}{*}{\noindent\begin{tabular}{p{1.5cm}}Region Discretization \end{tabular}} & Scale effect & Scale dependency & CA-\cite{damm2020we}, PA-\cite{RN323,zeng2020revisiting}, OD-\cite{gutierrez2020multi}, PM-\cite{yuan2020deep,sadri2018will}, V-\cite{RN323}, VA-\cite{malik2014proactive,zeng2020revisiting} &  PS-\cite{RN326}, TM-\cite{ling2020quality}, EA-\cite{RN305}, EP-\cite{RN337}, SA-\cite{RN334}, IoT-\cite{RN327}
\multirow{2}{*}{} \\ \cline{2-3}
 & Zoning effect & Zone dependency &   & \\ \hline

\multirow{11}{*}{\noindent\begin{tabular}{p{1.7cm}}Data Characteristics\end{tabular}} &

\multirow{2}{*}{\noindent\begin{tabular}{p{1.7cm}}Specificity \end{tabular} } & 
\begin{tabular}{p{2.7cm}}Spatiotemporal data tend to be unique to a particular space-time region.\end{tabular} & CA-\cite{rashidi2015spatial}, PM-\cite{zhao2015spatiotemporal}, V-\cite{RN66,van2012visualization}, VA-\cite{malik2014proactive}& PS-\cite{malik2014proactive}, EA-\cite{rashidi2015spatial} SA-\cite{zhao2015spatiotemporal} \\ \cline{3-3} 
 &  & 
\noindent\begin{tabular}{p{2.7cm}}Learned model is specific to a particular spatiotemporal region.\end{tabular}  & &  \\ \cline{2-5}
 
 & Vagueness & Similarities rooted from different criteria &  S-\cite{Shekhar2015perspective}, CA-\cite{RN307,kang2020urban}, PA-\cite{RN306}, OD-\cite{RN308}, PM-\cite{RN281} & PS-\cite{RN139,RN349}, TM-\cite{RN342}, EA-\cite{RN350}, EP-\cite{RN339}, SA-\cite{RN340}, IoT-\cite{RN340} \\ \cline{2-5}
 & Dynamic-ity & Continuous change through space and time. & CA-\cite{RN233}, PA-\cite{RN305}, OD-\cite{RN304,ferreira2020spatiotemporal}, PM-\cite{hens2019spatiotemporal,RN303,rumi2018theft}, V-\cite{cheng2020reading}  & PS-\cite{RN327}, TM-\cite{RN341,RN342}, EA-\cite{RN305}, EP-\cite{RN338}, SA-\cite{RN334}, IoT-\cite{RN327} \\ \cline{2-5}
 & Social & \noindent\begin{tabular}{p{2.2cm}}Correlation with the socio-economic characteristics\end{tabular} & CA-\cite{steiger2016exploration}, OD-\cite{chae2012spatiotemporal}, PM-\cite{Rumi2018Crime,zhao2015spatiotemporal,bogomolov2014once,RN116}, V-\cite{hochman2012visualizing,RN347} VA-\cite{RN334} & PS-\cite{bogomolov2014once}, SA-\cite{Rumi2018Crime,zhao2015spatiotemporal}, IoT-\cite{wang2020analysis,kaur2018shopping,chae2012spatiotemporal} \\ \cline{2-3}
  
 & \multirow{2}{*}{\noindent\begin{tabular}{p{1.7cm}} Network-ed \end{tabular}} & 
 Influence between objects and trajectories.&  &  \\ \cline{3-3} 
 &  & Exponential number of relationships. &  &  \\ \cline{2-5}
 
 & \multirow{2}{*}{\noindent\begin{tabular}{p{1.7cm}} Heterog-eneous and Non-stationary \end{tabular}} & 
\noindent\begin{tabular}{p{2.7cm}}Wide variation of data distributions over space and time.\end{tabular} & S-\cite{RN171,RN199}, CA-\cite{RN318}, PA-\cite{ren2018location,RN317}, OD-\cite{RN319}, PM-\cite{RN237,rahaman2018wait}, V-\cite{RN316} VA-\cite{malik2014proactive} & PS-\cite{kadar2019public,RN325}, TM-\cite{RN341,RN317}, EA-\cite{RN351}, EP \cite{RN331}, SA-\cite{RN334}, IoT-\cite{yang2019deep,RN353} \\ \cline{3-3} 
 &  & 
\noindent\begin{tabular}{p{2.7cm}} Different learning models for varying spatiotemporal regions. \end{tabular} & &  \\ \cline{2-5}
 
& Limited Access and Privacy & Privacy issues & S-\cite{RN151}, CA-\cite{acs2014case}, PA-\cite{de2013unique}, PM-\cite{huang2020spatiotemporal} & PS-\cite{ratcliffe2010crime}, TM-\cite{kaltenbrunner2010urban}, IoT-\cite{de2013unique} \\ \cline{2-5}
& Poor Quality & \noindent\begin{tabular}{p{2.7cm}} Uncertainties, Partial knowledge, Conjectures \end{tabular}
&  CA-\cite{RN139}, PM-\cite{diehl2015visual} V-\cite{islam2018uncertainty} VA-\cite{malik2014proactive} & PS-\cite{malik2014proactive,murray2011hybrid}, EA-\cite{diehl2015visual}, SA-\cite{steiger2016exploration}, IoT-\cite{wang2018extracting} \\ \cline{2-5}
& Big data & Volume, variety and velocity & CA-\cite{RN287,shao2016clustering}, PA-\cite{cheng2020high},PM-\cite{RN291}, OD-\cite{RN288}, V-\cite{RN293}& PS-\cite{RN256}, TM-\cite{RN345}, EA-\cite{RN350}, EP-\cite{RN337}, SA-\cite{RN347}, IoT-\cite{RN345} \\ \hline

\multirow{5}{*}{\noindent\begin{tabular}{p{1.7cm}}Open Issues\end{tabular}} &

 Data Representations &
 Limited representations of spatiotemporal data & 
 \cite{Santos2016representation,dunkel2019conceptual,gao2020generative,yu2020extracting,golany2020simgans} 
 &
 \\ \cline{2-5}
 &

 Advanced Modelling &
 Depend on high-density locations while ignoring the temporally related attributes. &  
 \cite{RN205,RN279,RN138,RN291,RN287,NEURIPS2019_455cb265,NEURIPS2018_69386f6b,NIPS2019_8730,kim2018spatio}
 &
 \\ \cline{2-5}&

 Visualisation &
 Developing techniques for spatial visualisation, while less consideration is given to spatiotemporal &  
 \cite{RN209,kastner2020visualizing,sakaue2020active,rizwan2020visualization,salcedo2020novel,sha2020spatiotemporal}
&
 \\ \cline{2-5}&

 Comprehen-sive Approaches &
 Focusing on certain problems and do not introduce comprehensive spatiotemporal solutions  & 
\cite{RN280}
&
 \\ \cline{2-5}&

 Fairness, Accountability, Transparency, and Ethics (FATE) &
 Amplifying genders, denying people services, and racial biases . &  
\cite{dudk2020assessing,olteanu2020when,buolamwini2018gender,raghavan2020mitigating,blodgett2020language,guo2019toward,bird2020fairlearn}
&
 \\ \hline
\end{longtable}
}

%
%
%

\section{Conclusion}\label{conclusion}
STDM is important due to the availability of large amounts of geographic and time-stamped data that can be mined to solve many interesting problems in different applications. STDM aims at discovering beneficial relationships and patterns that are implicit in spatiotemporal data. In this regard, STDM focuses on the design of efficient and scalable algorithms to mine, i.e., extract, predict, cluster, and quantify spatiotemporal patterns. Performing STDM is more difficult than traditional data mining due to the complex types of spatiotemporal relationships, interdisciplinary nature of data and tasks, and the unique characteristics of spatiotemporal data. 
Future research should focus on developing new modelling and visualisation methods that enable the integration of multiple STDM tasks to solve more complex scenarios. 
In this paper, we described the STDM problems and open gaps. We explained general issues related to spatiotemporal relationships, interdisciplinarity, discretisation, data characteristics, and research limitations. In an attempt to produce a comprehensive survey about the STDM challenges, we discussed the tasks and applications related challenges.
\section*{acknowledgements}
Ali Hamdi is supported by RMIT Research Stipend Scholarship. This work was funded by NPRP8-408-2-172 grant from Qatar National Research Fund (a member of Qatar Foundation). The statements made herein are solely the responsibility of the authors.

\bibliographystyle{spmpsci}   

\bibliography{STDM}


\end{document}